\documentclass[screen, authorversion]{acmart}

\usepackage{booktabs}
\usepackage{makecell}
\usepackage[shortcuts,acronym]{glossaries}
\usepackage{subfig}
\usepackage{tabularx}
\usepackage[]{mdframed}

\AtBeginDocument{%
  }

\setcopyright{acmlicensed}
\copyrightyear{2025}
\acmYear{2025}
\acmDOI{10.1145/3745771}

\acmJournal{TOCHI}
\acmVolume{32}
\acmNumber{6}
\acmArticle{59}
\acmMonth{8}

\newacronym{AI}{AI}{artificial intelligence}
\newacronym{AutoML}{AutoML}{automated ML}
\newacronym{HPO}{HPO}{hyperparameter optimization}
\newacronym{ML}{ML}{machine learning}
\newacronym{HCI}{HCI}{human-computer interaction}

\newcommand{\ie}{i.e., }
\newcommand{\eg}{e.g., }

\newcommand{\myquote}[2]{
    \begingroup
    \vspace{-2ex}
    \righthyphenmin=100
    ~\\``\textit{#2}''\newline\hspace*{\fill} --#1\\
    \vspace{-2ex}
    \endgroup
}

\begin{document}

\title{Practitioner Motives to Use Different Hyperparameter Optimization Methods}

\author{Niclas~Kannengießer}
\affiliation{%
  \institution{Karlsruhe Institute of Technology}
  \city{Karlsruhe}
  \country{Germany}
}
\email{niclas.kannengiesser@kit.edu}

\author{Niklas~Hasebrook}
\email{niklas.hasebrook@gmail.com}
\affiliation{%
  \institution{zeb consulting}
  \city{Berlin}
  \country{Germany}
}

\author{Felix~Morsbach}
\affiliation{%
  \institution{Karlsruhe Institute of Technology}
  \city{Karlsruhe}
  \country{Germany}
}
\email{felix.morsbach@kit.edu}

\author{Marc-Andr\'e~Z\"oller}
\affiliation{%
  \institution{University of Stuttgart}
  \city{Stuttgart}
  \country{Germany}
}
\email{mazoeller@gmail.com}

\author{Jörg~K.H.~Franke}
\affiliation{%
 \institution{University of Freiburg}
  \city{Freiburg}
  \country{Germany}
}
\email{frankej@cs.uni-freiburg.de}

\author{Marius~Lindauer}
\affiliation{%
  \institution{University of Hannover}
  \city{Hannover}
  \country{Germany}
}
\email{m.lindauer@ai.uni-hannover.de}

\author{Frank~Hutter}
\affiliation{%
 \institution{University of Freiburg}
  \city{Freiburg}
  \country{Germany}
}
\email{fh@cs.uni-freiburg.de}

\author{Ali~Sunyaev}
\affiliation{%
  \institution{Technical University of Munich (Campus Heilbronn)}
  \city{Heilbronn}
  \country{Germany}}
\email{sunyaev@tum.de}

\renewcommand{\shortauthors}{Kannengie{\ss}er et al.}

\begin{abstract}
Programmatic \ac{HPO} methods, such as Bayesian optimization and evolutionary algorithms, are highly sample-efficient in identifying optimal hyperparameter configurations for \ac{ML} models. However, practitioners frequently use less efficient methods, such as grid search, which can lead to under-optimized models.
We suspect this behavior is driven by a range of practitioner-specific motives. Practitioner motives, however, still need to be clarified to enhance user-centered development of \ac{HPO} tools.
To uncover practitioner motives to use different \ac{HPO} methods, we conducted 20 semi-structured interviews and an online survey with 49 \ac{ML} experts.
By presenting main goals (\eg~increase \ac{ML} model understanding) and contextual factors affecting practitioners' selection of \ac{HPO} methods (\eg available computer resources), this study offers a conceptual foundation to better understand why practitioners use different \ac{HPO} methods, supporting development of more user-centered and context-adaptive \ac{HPO} tools in \acl{AutoML}.
\end{abstract}


  
\received{25 September 2024}
\received[revised]{12 March 2025}
\received[accepted]{5 June 2025}

\maketitle

\section{Introduction}

The performance of \acf{ML} models is highly sensitive to their hyperparameter configurations~\cite{BergstraJames2012,melis2018state,Henderson2018,chen2018bayesian,zhang-aistats21,kadra2021welltuned}.
Identifying optimal hyperparameter configurations for training \ac{ML} models is, however, a complex and often daunting task---even for seasoned \ac{ML} experts---because of large search spaces of hyperparameter values and commonly unknown relationships between \ac{ML} model performance, hyperparameter configuration, and dataset.
Consequently, practitioners---individuals regularly involved in the development of viable \ac{ML} models that are meant for productive use in research or industry---experiment with various hyperparameter configurations to identify optimal ones, often relying on trial and error.
This iterative process of exploring, testing, and adjusting hyperparameter configurations related to \ac{ML} models is known as \acf{HPO}.

\ac{HPO} by manual tuning is often cumbersome, tedious, and error-prone.
To help practitioners with \ac{HPO}, research with a technical focus (\eg \cite{Hutter2018, BischlBLPRCTUBBDL23}) developed several programmatic \ac{HPO} methods, including grid search, random search, Bayesian optimization, and evolutionary algorithms, and implemented such methods as software tools (\eg Hyperopt and Hyperband; \cite{Bergstra2013, Li2018}).
Existing \ac{HPO} methods, programmatic and non-programmatic ones, differ considerably in the way they optimize hyperparameter configurations. Such differences lead some \ac{HPO} methods to superiority over others. That superiority is usually demonstrated using conventional performance metrics from computer science, including minimization of generalization errors and increase of sample efficiency \cite{Zoller2021,gijsbers2022amlb, Lindauer2022}. 
However, practitioners often use \ac{HPO} methods that are inferior according to conventional performance metrics \cite{bouthillier_survey_2020}. 
For example, practitioners often prefer to perform grid search over the more sample-efficient Bayesian optimization~\cite{Snoek2012}.

The dominant use of seemingly inferior \ac{HPO} methods suggests that practitioners have motives beyond the fulfillment of conventional goals pursued in \ac{HPO}, such as increasing \ac{ML} model performance.
However, the motives behind practitioners' choices of HPO methods---shaped by contextual factors---remain unclear.
This unclarity about why practitioners use which \ac{HPO} methods inhibits more user-centered development of \ac{HPO} methods and tools that support practitioners in attaining their goals, especially those beyond conventional performance metrics.
To support development of more user-centered \ac{HPO} methods and tools for \acf{AutoML} \cite{Vanschoren2019}, practitioner motives for \ac{HPO} need to be better understood. 
We approach the following research question: \emph{Why do practitioners choose different \ac{HPO} methods?}

We applied a two-step research approach consisting of an interview study and a survey study based on an online questionnaire.
First, we conducted semi-structured interviews with 20 \ac{ML} experts to unveil \ac{HPO} methods that practitioners commonly use, the goals pursued by practitioners when applying \ac{HPO} methods, and the contextual factors that influence practitioners' choices of \ac{HPO} methods to attain goals.
Second, we performed an online survey with 49 participants to collect evidence for the external validity of the relevance of the \ac{HPO} methods, goals, and contextual factors identified in the interviews.

Our main ambition is to support more user-centric development of \ac{AutoML} methods and tools for \ac{HPO} by bridging the gap between technical advancements and human factors. By integrating practitioner motives into \ac{HPO} research, we aim to ensure that future tools are not only technically sound but also aligned with real-world needs and decision-making processes.
In particular, this work has three main contributions.
First, this work presents a conceptual foundation constituted of principal goals (\eg improving \ac{ML} model performance and target audience compliance) and contextual factors (\eg compute resources and method traceability) influencing selections of \ac{HPO} methods. This is useful to enhance human-in-the-loop ML by clarifying information needs and improving practitioner engagement. Moreover, the conceptual foundation supports goal-driven development of \ac{HPO} methods and tools and new benchmarks with a focus on human factors.
Second, we present a mapping of HPO methods, goals, and contextual factors to explain practitioners' decision-making. This mapping informs user-centered HPO design and adaptive automation features. Moreover, it highlights key input parameters for developing context-sensitive HPO tools.
Third, we analyzed practitioners’ perceived success with different HPO methods, revealing strengths and areas for improvement. These findings contribute to the development of more effective decision-support systems for \ac{HPO} method selection.
Moreover, the reported success rates help identify limitations of HPO methods, guiding development of better-tailored and more effective \ac{HPO} methods and tools.

The remainder of this work is structured into five sections. First, we describe the state of research on \ac{HPO} in section~\ref{sec:related-work}. Section~\ref{sec:methods} reports the approach we applied to answer the research question. Then, we present the results of this study, including four \ac{HPO} methods, six goals, and fourteen contextual factors in section~\ref{results}. In section~\ref{sec:discussion}, we discuss the principal findings of this work, explain the contributions of this study, describe possible threats to the validity of the results, and outline future research directions. We conclude with our main takeaways in section~\ref{sec:conclusion}.

\section{Background and Related Work}
\label{sec:related-work}

Driven largely by technical advancements, \ac{HPO} is a key research area with significant potential to shape \ac{ML} model development \cite{Hutter2018b}.
To support a better understanding of the results of this study, we briefly describe the technical foundations of principal \ac{HPO} methods. Moreover, we elucidate related research on involving human beings in automated \ac{HPO}, a prominent field within \ac{AutoML} research.

\subsection{Hyperparameter Optimization Methods}
\label{sec:related-work-hpo-methods}

There are five principal, model-agnostic \ac{HPO} methods: manual tuning, grid search, random search, evolutionary algorithms, and Bayesian optimization.\footnote{There are also gradient-based methods, but because these methods are often perceived as being brittle and are often model-specific, they are out of the scope of this study.} These \ac{HPO} methods are briefly described below.

\emph{Manual tuning} refers to a set of \ac{HPO} methods where practitioners choose hyperparameter configurations based on explicit and implicit knowledge and influences of contextual factors.
The dependency of manual tuning on practitioners' experiences and even unconscious heuristics in decisions \cite{Gigerenzer_Brighton_2009} makes manual tuning very individual to practitioners, rendering the explication of manual tuning difficult. Thus, manual tuning is usually hardly replicable~\cite{musgrave2020metric, simon2023hpusage}.
Commonly, only intermediate information (\eg~tuned hyperparameters) is available, while reasons for selecting hyperparameter configurations often remain unclear.
This complicates the formalization of specific \ac{HPO} methods in manual tuning, making it unclear how practitioners actually proceed.
Common strategies include starting optimization from well-performing hyperparameter values \cite{xin2021wither}, using a sound experiment design for incremental improvement~\cite{tuningplaybookgithub}, and removing irrelevant hyperparameters from the search space~\cite{Radosavovic_2020_CVPR}.
In addition, difficulty in explication make it hard to evaluate the sample efficiency of strategies in manual tuning.
While evidence that manual tuning outperforms advanced \ac{HPO} methods is lacking, prior publications offer initial evidence that advanced \ac{HPO} methods can outperform manual tuning in several use cases~\cite{feurer_2016,melis2018state,chen2018bayesian,zhang-aistats21,kadra2021welltuned,Zoller2023}.

To tackle shortcomings of manual tuning, \ac{AutoML} \cite{Hutter2018} is envisioned to automate all aspects related to development of \ac{ML} models in a problem-agnostic manner, for example, by programmatic \ac{HPO} methods.
Research on \ac{AutoML} has primarily approached \ac{HPO} from a technical perspective (\eg \cite{Hutter2018,BischlBLPRCTUBBDL23}).
Typical works (\eg \cite{Bergstra2011, Jamieson2016, Golovin2017, Falkner2018, Li2018, He2018, Wang2021b}) focus on performance optimization of \ac{ML} models in terms of smaller generalization errors, smaller \ac{ML} model size, or lower latency.
%
Often, such works investigate \ac{HPO} from a mathematical perspective and treat \ac{HPO} as a black-box optimization problem:
given a problem instance in the form of a dataset and a loss function, a black-box optimizer (\eg Bayesian optimization) searches for hyperparameter configurations in a predefined search space to enhance an \ac{ML} model in terms of a given metric (\eg accuracy on a validation set).
Multi-objective optimization methods can be used to specify additional properties of the resulting \ac{ML} model, such as algorithmic fairness, fast inference, and low model complexity~\cite{Gardner_Golovidov_Griffin_Koch_2019,Binder_Moosbauer_Thomas_Bischl_2020,karl2022multi,dooley2022on}.

\emph{Grid search} is one of the earliest \ac{HPO} methods that can be executed programmatically to solve the black-box optimization problem.
Grid search refers to the process of evaluating the Cartesian product of a finite set of hyperparameter configurations.
Every possible combination of hyperparameter values included in the defined subset of the search space is evaluated~\cite{montgomery2017design}. Thus, grid search does not scale well with the number of hyperparameters.
Grid search relies on a deterministic procedure to select hyperparameter configurations to be evaluated. The deterministic procedure allows reproducing experiments.
For reproduction, the originally applied search space and discretization strategy must be known.

Although easy to implement, parallelize, and reproduce, grid search has become increasingly unsuited for modern \ac{HPO} problems due to the curse of dimensionality~\cite{BergstraJames2012}.
In practice, not all hyperparameters have a similar impact on the final performance of \ac{ML} models~\cite{BergstraJames2012,VanRijn2018}.
Due to its rigid search strategy, grid search often allocates much of the optimization budget to less relevant regions of the search space~\cite{BergstraJames2012}.
Put differently, sample efficiency of grid search tends to be lower compared to later \ac{HPO} methods (\eg random search, Bayesian optimization, and evolutionary algorithms~\cite{Snoek2012,Eggensperger2013,turner_bayesian_2021}), in particular, because grid search cannot make use of the low effective dimensionality of \ac{HPO} problems~\cite{Bergstra2011}.

\emph{Random search} refers to the process of sampling random hyperparameter configurations from a defined search space until a budget is exhausted~\cite{BergstraJames2012}.
Random search handles hyperparameters of varying importance more effectively than grid search and achieves greater sample efficiency in high-dimensional search spaces, especially when hyperparameters have differing influence on \ac{ML} model performance~\cite{BergstraJames2012}.
Random search can be reproduced if the used search space, the randomness generator, and the corresponding seed are known.

To address large search spaces, grid search and random search have been supplemented by methods that exploit knowledge of well-performing regions within the set of possible hyperparameter configurations \cite{Snoek2012}.
A well-established approach for balancing exploration and exploitation is evolutionary optimization.
Inspired by biological evolution, \emph{evolutionary algorithms} iteratively mutate a population of candidate solutions to obtain solutions with better performance.
Evolutionary algorithms often perform well in optimizing black-box functions \cite{Olson2016} but are rather inefficient in terms of samples \cite{BischlBLPRCTUBBDL23}.
\ac{HPO} based on evolutionary algorithms can be reproduced with fixed random seeds if the search space and randomness generator are known.

\emph{Bayesian optimization} can be an alternative to evolutionary algorithms.
Bayesian optimization refers to the process of using a sequential approach based on a surrogate model to find appropriate hyperparameter configurations for \ac{ML} models in defined search spaces (\eg \cite{Brochu2010,Shahriari2016,Hutter2018,garnett_bayesoptbook_2022}).
In Bayesian optimization, an optimizer constructs an internal probabilistic model, mapping hyperparameter configurations to expected \ac{ML} model performance, to achieve an optimal balance between exploration and exploitation \cite{Shahriari2016,Frazier2018, garnett_bayesoptbook_2022}. Bayesian optimization can be extended to deal with high-dimensional search spaces through, for example, additive surrogate models~\cite{KandasamyVNPC0P20} or local trust regions~\cite{ErikssonPGTP19}.
\ac{HPO} based on Bayesian optimization can be reproduced with fixed random seeds if the search space, the acquisition function, and the surrogate model, including its hyperparameters, are known.

\subsection{Research on Hyperparameter Optimization}
\label{sec:related-work-hpo-methods-research}

Extant research on \ac{HPO} methods is mainly driven by technological advancements that aim at increasing sample efficiency of \ac{HPO} methods \cite{Bergstra2011, Hutter2011, Snoek2012}, reducing time for evaluating objective functions~\cite{Swersky2014, Domhan2015, Li2018}, and transferring knowledge from prior optimization runs to similar problem instances~\cite{8955514, Vanschoren2019}. Such technological advancements are valuable but designed to reach better results in terms of conventional performance metrics. Practitioner motives to use \ac{HPO} methods beyond conventional performance metrics are largely neglected.

Studies exploring practitioners’ experiences with programmatic \ac{HPO} methods provide valuable insights into the perceived advantages and disadvantages of these methods~\cite{Gil2019a, Wang2019a, Wang2021}.
Practitioners acknowledge advantages of programmatic \ac{HPO} methods, which are commonly related to faster turn-around time for building \ac{ML} models and, thus, higher productivity in developing \ac{ML} models \cite{Wang2019a, Wang2021a}.
Automatically tuned \ac{ML} models are often used by practitioners to create initial baseline \ac{ML} models for subsequent manual tuning or to gain data insights~\cite{Wang2019a,Wang2021a,Crisan2021}.
However, \ac{AutoML} practitioners often bemoan insufficient confidence in the results of programmatic \ac{HPO} methods and, therefore, refuse to blindly use \ac{ML} models optimized with programmatic \ac{HPO} methods \cite{lee2020human, Wang2019a, Drozdal2020, Khuat2022}.
Even though acknowledging that programmatic \ac{HPO} is useful to develop well-performing \ac{ML} models \cite{Wang2021a}, practitioners often refuse to use those \ac{HPO} methods to not be accountable for \ac{ML} models they do not understand~\cite{Drozdal2020}.
A lack of confidence is often linked to the perceived black-box nature of programmatic \ac{HPO} methods, such as Bayesian optimization and evolutionary algorithms, limiting transparency of optimizer internals.
Practitioners prefer support that augments their daily data science work (\eg through guidance) rather than fully automating it \cite{Crisan2021}.

Research on human-guided \ac{HPO} focuses on involving humans in programmatic \ac{HPO} methods to improve \ac{HPO} with dormant domain expertise \cite{Wang2019a, higuchi2021interactivehpo, wang2019atms}.
This requires identifying how and when to involve humans in \ac{HPO} to achieve the best combination of automation and human knowledge \cite{Crisan2021}.
Especially involvement of practitioners in \ac{ML} model development, including \ac{HPO}, seems promising for a higher level of automation.
For other tasks, including data acquisition and requirement analysis, practitioners prefer strong human involvement with a low level of automation \cite{Wang2021}.
Interactions of practitioners with software tools for programmatic \ac{HPO} were structured into different modes of cooperation between practitioners and software tools, ranging from manual tuning to full automation, in the literature~\cite{lee2020human, Crisan2021, Wang2021}.

Extant research describes valuable concepts how practitioners could interact with software tools for programmatic \ac{HPO} \cite{sun2023automl, xin2021wither} and how to design visual analytics tools for \ac{HPO} to support practitioners \cite{higuchi2021interactivehpo, Zoller2022, wang2019atms}.
Yet, the different programmatic \ac{HPO} methods are not further differentiated, and practitioner motives to select different \ac{HPO} methods remain unclear.
Supporting better understanding of practitioner motives for selecting \ac{HPO} methods is the main goal of this study.

\section{Methods}
\label{sec:methods}

We applied a mixed-methods research approach consisting of two main steps. First, we conducted semi-structured interviews with \ac{ML} experts to develop a set of commonly used \ac{HPO} methods, goals pursued by using \ac{HPO} methods, and contextual factors that influence the choice for \ac{HPO} methods. Second, we conducted a survey using an online questionnaire to collect evidence of the external validity of the interviews. The following details the two steps.

\subsection{Semi-structured Interviews with Machine Learning Experts}
\label{subsec:expert-interviews}

To identify practitioners' goals pursued in \ac{HPO} and understand decisions for specific \ac{HPO} methods to achieve these goals, we chose an exploratory, qualitative research approach and conducted semi-structured expert interviews.

\paragraph{Data Collection}
To find interviewees for the study, we reached out to personal contacts from ongoing research projects, authors of scientific studies, and companies that develop \ac{ML} models.
The contacted persons had heterogeneous experiences with \ac{HPO} and \ac{ML}, ranging from novices to experts and different \ac{ML} fields (\eg~computer vision, natural language processing, and reinforcement learning).
Among the contacted potential interviewees, 20 agreed to participate in the interview study.

The interviewed experts were all \ac{ML} practitioners---individuals who regularly develop ML models for practical use in research or industry. 
All interviewees had actively contributed to at least one successfully deployed \ac{ML} model in research or industry.
The experts were associated with thirteen different organizations and had an average work experience in \ac{ML} of about five years (see Table~\ref{tab:interviewees}).
Among the interviewees, two held a PhD and 12 were PhD students from academia with Master's degrees---most nearing completion of their doctorates.
All industry participants held at least a Master's degree and had, on average, more than five years of experience developing \ac{ML} models for production.
The 12 PhD students worked at universities in the field of applied \ac{ML}. One PhD worked in the industry as an \ac{ML} engineer, and another worked at a university in the field of applied \ac{ML}. Two interviewees with Bachelor’s degrees were pursuing Master's degrees while working as ML developers in industry as student trainees. 

Ten participants had a skill level of \ac{ML} innovators---practitioners engaged in ML research, including studies on core algorithms and the application of ML in scientific discoveries \cite{xin2021wither}---while the other ten were ML engineers---experienced ML practitioners with formal training in applied \ac{ML} \cite{xin2021wither}.
The study participants used \ac{HPO} in the context of computer vision, natural language processing, reinforcement learning, and time series forecasting.
Interviewees from the industry also mentioned to have used these \ac{ML} techniques in bioinformatics~(2), robotics~(2), e-commerce (1), and finance~(1).
Because \ac{AutoML}, including \ac{HPO} based on programmatic methods, aims to be domain agnostic \cite{Hutter2018}, usage of \ac{AutoML} tools should be independent of the different domains.

\begin{table}[b]
    \renewcommand{\arraystretch}{1.25}
    \centering
    
    \caption{Overview of the demographic data of the 20 interviewees. The numbers in parentheses show the number of interviewees with the respective characteristics. The interviewees could name multiple \ac{ML} fields.}
    \label{tab:interviewees}
    
    \begin{tabularx}{\linewidth}{p{2.0cm}p{2.8cm}p{2.8cm}p{2.7cm}X}
    \toprule
    Field & Highest Degree of \newline Education & Years of Experience & Skill Level & \ac{ML} Field\\
    \midrule
     Academia~$(14)$  & Bachelor~$(2)$    & $< 2$~$(4)$       & \ac{ML} Innovator (10) & CV~$(8)$ \\
     Industry~$(6)$   & Master~$(16)$     & $2$--$4$~$(6)$    & \ac{ML} Engineer (10)  & NLP~$(6)$ \\
                      & PhD~$(2)$         & $5$--$7$~$(7)$    &                    & RL~$(5)$ \\
                      &                   & $> 7$~$(3)$       &                    & TSF~$(3)$ \\
                      
    \bottomrule
    \multicolumn{5}{l}{\footnotesize \em CV: Computer vision, NLP: Natural language processing, RL: Reinforcement learning, TSF: Time series forecasting}\\
    \end{tabularx}
\end{table}

We developed an interview guide for the semi-structured interviews \cite{myers2007qualitative,gorden1975interviewing}. The interview guide structured the interviews into four sections: \textit{briefing}, \textit{\ac{HPO} in \ac{ML}}, \textit{participant background and personal experiences}, and \textit{debriefing}.
Each section outlined its purpose and the corresponding interview questions.
We sent the interview guide to participants in advance to help them prepare for the interview.
Since the interviews focused on a past \ac{ML} project, we asked the participants to choose one to discuss in their interviews.
As we also recruited a few participants from personal contacts, we aimed to avoid social desirability bias; we ensured that no interviewee was acquainted with the interviewer.
Moreover, we phrased the questions as neutral and open-ended to encourage unbiased and detailed responses.
For each interview, we created a non-judgemental atmosphere and clarified that there are only valid answers to our questions with valuable insights important to our research. We also explained to the interviewees that their responses would be anonymized and no links to their identity would be possible.

We started each interview with a briefing of the participants by explaining the background and goals of the study.
Second, in line with the section \textit{\ac{HPO} in \ac{ML}} in the interview guide, we asked the \ac{ML} experts to name \ac{HPO} methods they used in \ac{ML} projects.
Moreover, we gathered insights into why they selected \ac{HPO} methods for optimization and asked about contextual factors that influenced their selection of \ac{HPO} methods.
Third, we wanted to learn more about the interviewees' \textit{participant background and personal experiences}, including years of experience in \ac{ML} development, main areas of using \ac{ML}, and highest degree of education.
Fourth, in \textit{debriefing}, we asked the interviewees for their final thoughts and remarks on the study and topic and informed them about the further proceeding in the context of the study.
We conducted interviews with \ac{ML} experts following methodological guidelines, ensuring an open-minded approach without pressure on the interviewee and avoiding any influence on their responses through neutral questions~\cite{gorden1975interviewing, Barriball_While_1994, McIntosh_Morse_2015}.
The 20 interviews took between 18 and 61 minutes, with an average time of 31 minutes.
We transcribed the interviews in preparation for the analysis.

\paragraph{Data Analysis}

We analyzed each interview transcript using thematic analysis~\cite{Braun_Clarke_2006, Braun_Clarke_2012} in groups of three authors.
Thematic analysis comprises six steps: (1) familiarize yourself with the data, (2) generate initial codes, (3) search for themes, (4) review themes, (5) define and name themes, and (6) produce the report.

Three of the authors coded each interview together. After familiarizing themselves with the transcripts (Step~1), the authors independently coded them (Step~2) to identify \ac{HPO} methods applied by practitioners, to extract practitioners' goals in \ac{HPO}, and to reveal contextual factors that influence the interviewees' decisions for \ac{HPO} methods.
We incorporated contextual factors to better understand influences on practitioners' decisions for using \ac{HPO} methods.
The three authors independently read the transcripts, identified quotes relevant to this study, and labeled the quote with a name (\ie a code) that expresses a potentially relevant \ac{HPO} method, goal, or contextual factor.
Each of the three authors coded the transcripts independently without a predefined coding scheme.
After coding each interview independently, the authors presented their coding results to each other and discussed and harmonized their results in multiple iterations per interview and across interviews.
In each iteration, the analysts aimed for mutually exclusive codes while ensuring comprehensive coverage.
Through this explorative approach involving three analysts per interview, we aimed to reduce biases arising opinions of single analysts.
The first coding iteration revealed 241 preliminary codes.

The authors harmonized their codes into 21 mutually exclusive codes so that no different codes had the same semantics.
For example, the coders merged the contextual factors \textit{knowledge about Bayesian optimization} and \textit{knowledge about grid search} into the contextual factor \textit{\ac{HPO} method comprehension}.
During the harmonization process, the authors aimed for unambiguous agreements regarding the codes and their intended meaning. To achieve this goal, the authors resolved conflicts in their coding results through intensive discussion and refinement.

Three of the authors developed candidate themes (Step~3) to group the harmonized codes based on semantic relationships.
If a code did not suit an existing theme, we created a new theme.
For example, we assigned the contextual factor \textit{available compute resources} to the theme \textit{technical environment}, while we created a new theme \textit{own knowledge} for the contextual factor \textit{\ac{HPO} method comprehension}.
The set of candidate themes was comprised of four themes associated with \ac{HPO} methods, six themes associated with goals, and three themes associated with contextual factors.

In Step~4, we reviewed and refined the candidate themes within the author team in multiple iterations. We again aimed to reach mutual exclusiveness of the themes and assigned codes and maintain exhaustiveness of the results.
Subsequently, we developed an intuitive name for each theme and a definition (Step~5).
Finally, we assigned the set of 13 themes to three categories: \ac{HPO} methods, principal goals, and contextual factors, and wrote up a summary of the results (Step~6).

After coding the transcripts of all 20 interviews, the analysis of the last 7 transcripts did not reveal additional \ac{HPO} methods, goals, and contextual factors. We assume to have reached theoretical saturation \cite{fusch2015saturation,guest2006saturation} after the first 13 interviews with 7 interviews confirming this assumption.
These last seven interviews analyzed were from three practitioners from industry and four from academia who hold a master's degree and are pursuing a PhD in the field of applied machine learning.
Given the diverse backgrounds of the interviewees and the substantial number of confirmatory interviews, we considered the results sufficiently robust and moved on with an online survey to collect evidence for the external validity of the coding results.

\subsection{Online Survey}
\label{subsec:questionnaire}
We conducted a survey study using an online questionnaire to collect evidence for the external validity of the interview study results and to learn whether practitioners perceive that they succeeded in achieving their goals through their decisions to use specific \ac{HPO} methods.

\paragraph{Questionnaire Structure}
The online questionnaire was structured into four sections: \textit{Introduction}, \textit{\ac{HPO} Methods and Goals}, \textit{Contextual Factor Integration}, and \textit{Demographics}.
In the \emph{Introduction} section, we described the motivation for and the structure of the questionnaire.
In \emph{Methods and Goals}, we showed participants a matrix that listed all goals and \ac{HPO} methods identified in the interview study.
Participants were asked to select all pairs of \ac{HPO} methods and goals to achieve specific goals.  Because the main purpose of the survey was to collect evidence for the validity of the results from the interview study, study participants could only select \ac{HPO} methods (\ie manual tuning, grid search, random search, Bayesian optimization, and evolutionary algorithms) and goals (\eg increase \ac{ML} model understanding, decrease necessary computations, and decrease practitioner effort) extracted from the previous interview study.
In a second question, we asked participants to indicate, for each selected pair of \ac{HPO} method and goal, whether they feel to have successfully achieved each goal.

In the \textit{Contextual Factor Integration} section, we wanted to better understand how practitioners perceived the influence of contextual factors identified in the interview study on their selections of \ac{HPO} methods.
For each pair of \ac{HPO} methods and goals previously selected, we asked the participants to express their perceived influence of each contextual factor on the selection of an \ac{HPO} method to achieve a particular goal on a five-point Likert scale---0 represents very low perceived influence, two corresponds to a neutral response (\ie the contextual factor was not perceived as influential), and four represents very high perceived influence.

Finally, in the \textit{Demographics} section, we collected information about the participants to provide context to the data collected.

\paragraph{Data Gathering}
To solicit participants for the online questionnaire, we contacted practitioners via email and promoted the study via social media platforms. We did not invite participants from the interview study.

In total, 166 participants filled out the first question of the \textit{\ac{HPO} Methods and Goals} section with 85 participants completing the whole section.
Only 57 participants also completed the subsequent \textit{Contextual Factor Integration} section. Of those, 8 participants did not provide demographic details, leading to a total of 49 participants completing the questionnaire.
Most of these 49 participants worked in large organizations with more than 500 employees, including automotive companies, companies specializing in IT support and services, and universities.
Table~\ref{tab:questionnaire-participants} shows more demographic details about the participants who completed the questionnaire.

\begin{table}[ht]
    \renewcommand{\arraystretch}{1.25}
    \centering

    \caption{Overview of the demographic data of the 49 participants that completed the online questionnaire. The numbers in parentheses show the number of interviewees with the respective characteristics. The participants could choose multiple \ac{ML} fields.}
    \label{tab:questionnaire-participants}
    
    \begin{tabularx}{\linewidth}{p{2.0cm}p{2.8cm}p{2.8cm}p{2.7cm}X}
    \toprule
    Field & Highest Degree of \newline Education & Years of Experience & Skill Level & \ac{ML} Field\\
    \midrule
     Academia~$(27)$    & High School~$(1)$ & $< 2$~$(8)$       & \ac{ML} Innovator (22) & CV~$(12)$\\
     Industry~$(22)$    & Bachelor~$(2)$    & $2$--$4$~$(20)$   & \ac{ML} Engineer (19) & TD~$(19)$\\
                        & Master~$(27)$     & $5$--$7$~$(14)$   & Novice (8)& NLP~$(15)$\\
                        & Diploma~$(2)$     & $8$--$10$~$(4)$   && TSF~$(8)$\\
                        & PhD~$(17)$        & $> 10$~$(3)$      && RL~$(5)$\\
    \bottomrule
    \multicolumn{5}{l}{\footnotesize \em CV: Computer vision, TD: Tabular data, NLP: Natural language processing, TSF: Time series forecasting, RL: Reinforcement learning}\\
    \end{tabularx}
\end{table}

\paragraph{Data Analysis}
By analyzing the responses to the online questionnaire, we sought to learn how frequently practitioners tend to choose which \ac{HPO} methods to pursue specific goals, given which contextual factors.
To prepare the analysis, we discarded all responses from participants who did not complete the questionnaire.
We only analyzed completed survey responses.
We extracted the number of identical responses and related them to each other.

\section{Results}
\label{results}
The study participants (\ie the interviewees and the survey participants) applied five principal \ac{HPO} methods to pursue six goals, influenced by fourteen contextual factors (see sections~\ref{results-hpo-methods}--\ref{subsec:results-dec-fac}).
The study participants pursued their goals with different self-reported success (see section ~\ref{subsec:results-perceived-success}).

\subsection{Hyperparameter Optimization Practices}
The following first briefly describes what \ac{HPO} methods the study participants used.
Second, we report the goals pursued by the study participants and which of the four \ac{HPO} methods practitioners used to attain what goals.
Third, we introduce fourteen contextual factors and how they can influence practitioners in their decisions for \ac{HPO} methods to achieve specific goals.

\subsubsection{Hyperparameter Optimization Methods Used by Practitioners}
\label{results-hpo-methods}

\begin{figure*}[ht]
    \centering
    \subfloat[Percentage of survey participants that already used the individual \ac{HPO} methods.]{\includegraphics[width=0.42\textwidth]{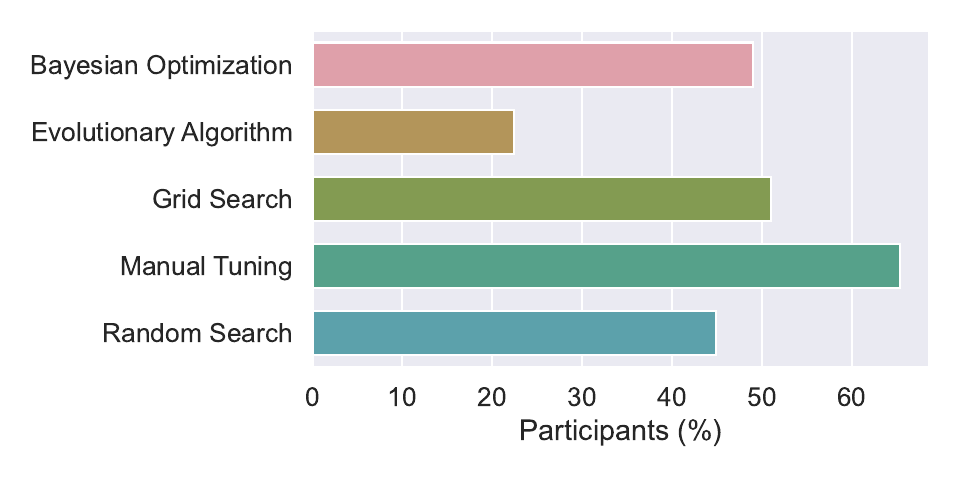}%
    \label{fig:hpo_method_distribution}}
    \qquad\qquad\qquad
    \subfloat[Number of \ac{HPO} methods used by participants.]{\includegraphics[width=0.42\textwidth]{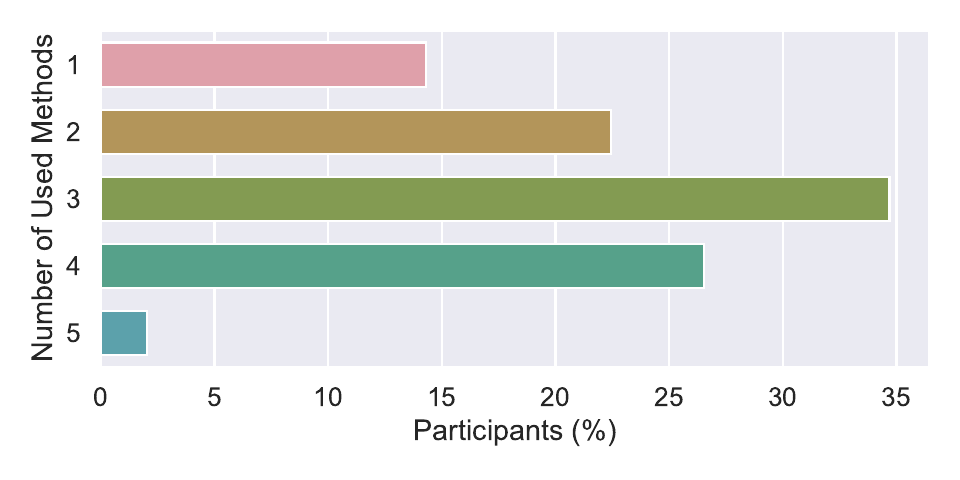}%
    \label{fig:multiple_hpo_method}}
    
    \caption{Overview of \ac{HPO} methods used by 49 study participants.}
    \label{fig:ovw-hpo-methods}
    \Description{Fully described in the text.}
\end{figure*}

The study participants applied five \ac{HPO} methods:
manual tuning, grid search, random search, Bayesian optimization, and evolutionary algorithms.
Each \ac{HPO} method was discussed by between two and eight interviewees.
Most interviewees discussed one or two \ac{HPO} methods in detail.\footnote{We asked the interviewees to discuss a prominent recent example of how they used \ac{HPO} methods, not to discuss all \ac{HPO} methods they ever used.} 
Most survey participants primarily used manual tuning, followed by grid search, Bayesian optimization, random search, and evolutionary algorithms (see Figure~\ref{fig:hpo_method_distribution}).
Academic participants tended to rely more on manual tuning and grid search compared to those from industry.
Most survey participants stated to have used at least three \ac{HPO} methods in their past \ac{ML} projects; about 2\% have even used at least five \ac{HPO} methods.
Only roughly 15\% of the survey participants have used only a single \ac{HPO} method (see Figure~\ref{fig:multiple_hpo_method}).

Even though literature indicates that Bayesian optimization yields better results in a shorter time than evolutionary algorithms, grid search, and random search (\eg \cite{Snoek2012,turner_bayesian_2021}), practitioners tend to use seemingly inferior \ac{HPO} methods.
Practitioners seem to not only aim at finding hyperparameter configurations for optimal \ac{ML} model performance but also pursue different goals.

\subsubsection{Goals of Practitioners Pursued with Different HPO Methods}
\label{subsec:results-method-goals}

We identified six goals that the participants pursued in \ac{HPO} (see~Table~\ref{tab:goals}). In the following, we first introduce each goal based on the interview results. Then, we describe the results of the survey study.

\begin{table*}[b]
    \renewcommand{\arraystretch}{1.25}
    \centering

    \caption{Principal goals practitioners pursue in \ac{HPO}}
    \label{tab:goals}
    
    \begin{tabularx}{\textwidth}{@{} lX @{}}
    \toprule
     Goal & Description \\
    \midrule
    
     \makecell[lt]{Comply with Target Audience} &
     The state where the applied \ac{HPO} method and the resulting \ac{ML} model fulfill the expectations of addressees\\
     
     \makecell[lt]{Decrease Necessary Computations} &
     The state where an \ac{ML} model is trained with an \ac{HPO} method that requires less compute resources than other methods but still is sufficiently useful for a given purpose\\

     \makecell[lt]{Decrease Practitioner Effort} &
     The state in which a practitioner applies an \ac{HPO} method for training an \ac{ML} model that requires less resources compared to other \ac{HPO} methods (\eg~time for learning a new \ac{HPO} method or implementing corresponding software tools)\\

     \makecell[lt]{Increase \ac{ML} Model Performance} &
     The state where a refined version of an \ac{ML} model outperforms its original version in terms of a specified metric\\

     \makecell[lt]{Increase \ac{ML} Model Understanding} &
     The state where a practitioner is able to predict changes in an \ac{ML} model's behavior caused by altering hyperparameter configurations based on an understanding of the \ac{ML} model's inner workings\\

     \makecell[lt]{Satisfy Requirements} &
     The state where the development and training of an \ac{ML} model satisfies social and technical demands imposed by stakeholders\\
    \bottomrule
    \end{tabularx}
\end{table*}

\paragraph{Comply with Target Audience}
\label{results-selection-tar-aud-com}
The goal \emph{comply with target audience} refers to aligning practitioners' choices with the expectations of their target audience.
Three interviewees from academia stated they had decided on \ac{HPO} methods to comply with the expectations of their target audiences regarding applied \ac{HPO} methods and the resulting \ac{ML} model.
For example, two of the three interviewees described Bayesian optimization as uncommon in their research communities and felt obliged to explain it in their scientific publications.
This would have required additional explanations of Bayesian optimization, even though the authors assumed the exact \ac{HPO} method was not relevant to their scientific work.
Therefore, they decided to use grid search as they assumed this \ac{HPO} method to be well-known in their research communities.

\paragraph{Decrease Necessary Computations}
\label{results-selection-dec-nec-com}

Extensive searches for optimal hyperparameter configurations in large search spaces typically require substantial computational resources.
Necessary computations for \ac{HPO} can be decreased by using an \ac{HPO} method that requires fewer compute resources than other methods but is still sufficiently useful.

\myquote{Interviewee~\#8, \ac{ML} Innovator in Academia}{This whole method was already super, super expensive [...] and if you would perform hyperparameter optimization again, then it becomes even more expensive.}

In contrast to the goal described before (\textit{comply with target audience}), decreasing necessary computations is not self-contained. Instead, it functions as a supplementary objective, generally addressed in conjunction with at least one goal and in response to contextual constraints. For example, practitioners may seek to decrease necessary computations to optimize the use of limited resources, enabling adequate improvements in \ac{ML} model performance.

\paragraph{Decrease Practitioner Effort}
\label{results-selection-dec-dev-effort}

Practitioners choose \ac{HPO} methods to reduce overhead, for example, in terms of the additional time required to understand the \ac{HPO} method or to integrate the \ac{HPO} method into workflows.
One industry interviewee reported that the sheer number of advanced \ac{HPO} method made it challenging to select the most suitable method for their use case. Experiencing the paradox of choice~\cite{schwartz2004paradox}, practitioners felt uncomfortable committing to a particular \ac{HPO} method and tool and therefore opted for manual tuning instead.

To decrease their efforts in \ac{HPO}, the interviewees applied grid search and manual tuning.
In particular, practitioners stated to have applied manual tuning to avoid efforts related to setting up \ac{HPO} tools in cluster infrastructures.

\myquote{Interviewee~\#5, \ac{ML} Innovator in Industry}{\ac{HPO} is time-consuming sometimes because it requires some extra lines of code to wrap all your models with this \ac{HPO} method and then set up the scripts to run them on a cluster.}

Like \textit{decrease necessary computations}, the goal \textit{decrease practitioner effort} is not self-contained, and commonly used in combination with other goals. For example, study participants used \ac{HPO} tools to enhance \ac{ML} model understanding by iteratively refining manually defined sets of hyperparameter values. Those practitioners leveraged \ac{HPO} tools to automate the reconfiguration of defined hyperparameter sets to accelerate manual tuning.

\paragraph{Increase \ac{ML} Model Understanding}
\label{results-selection-mod-com}
Increasing \ac{ML} model understanding refers to reaching the state where a practitioner can predict changes in an \ac{ML} model’s behavior due to tuning hyperparameter values based on an understanding of the \ac{ML} model's inner workings.
To increase their understanding of \ac{ML} models, the interviewees reported having applied manual tuning.
The interviewees claimed that manual tuning can improve their understanding of hyperparameter influences on \ac{ML} models as they formulate hypotheses about hyperparameter influences on \ac{ML} models and evaluate them immediately.
The interviewees explained that they iteratively improve their \ac{ML} model understanding by tuning hyperparameter values and testing their hypotheses.
This dynamic between \ac{ML} practitioner and \ac{ML} model is researched as interactive \ac{ML}. Interactive \ac{ML} supports practitioners in actively exploring the model space and testing hypotheses about hyperparameter effects \cite{Jiang2019}. In addition, visual analytics tools help practitioners by providing graphical representations of model behavior and parameter interactions (\eg~\cite{Spinner2019, Heyen2020}).

\paragraph{Increase \ac{ML} Model Performance}
\label{results-selection-imp-mod-per}
\ac{ML} model performance is increased when a refined version of the \ac{ML} model outperforms its original version in terms of a specified metric.
The interviewees chose manual tuning, grid search, random search, and Bayesian optimization \ac{HPO} methods to achieve this goal
for example, to prototype novel \ac{ML} models.

\myquote{Interviewee~\#6, \ac{ML} Engineer in Academia}{If the only concern is to find the best model possible and no one asks how I got there, and I do not have a lot of time, I probably would use a random search.}

\paragraph{Satisfy Requirements}
The goal to satisfy requirements refers to reaching the state in which the development and training of an \ac{ML} model fulfill social and technical constraints imposed by stakeholders (\eg~business clients, ethics commissions) and the environment (\eg~available compute resources).
Ten interviewees described that their decisions for \ac{HPO} methods were influenced by the goal of fulfilling such requirements.
For example, one interviewee reported preferring manual tuning to meet hard-to-formalize requirements, such as a smooth behavior of the model output.\\

\begin{figure}[ht]
\includegraphics[width=0.7\linewidth]{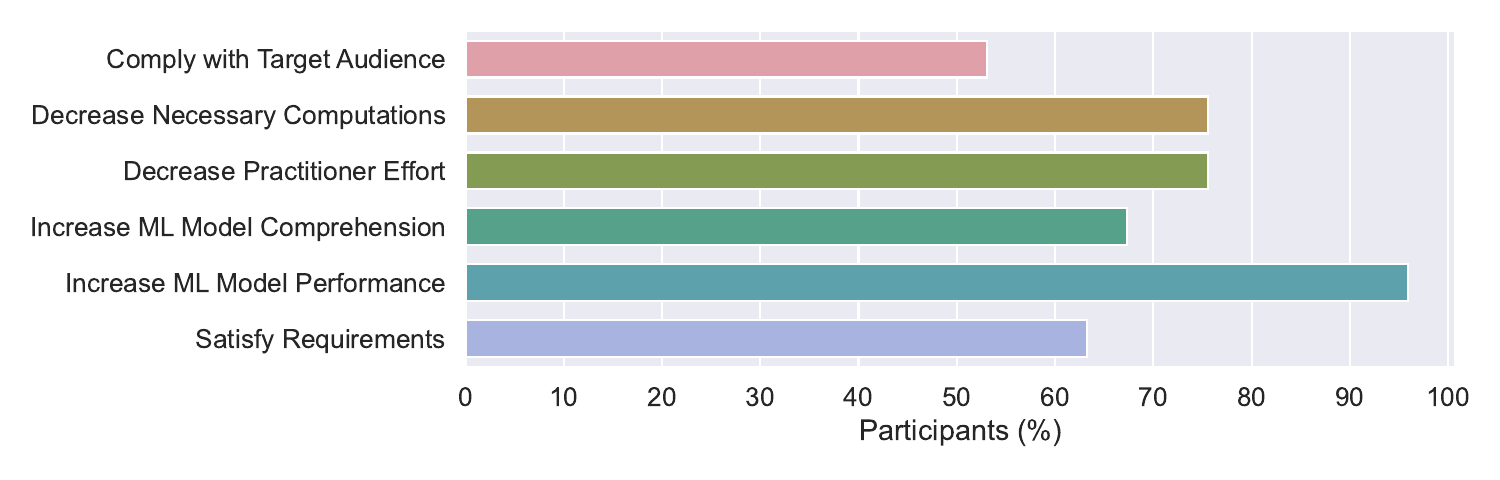}
\centering
\caption{Relative frequency of pursued goals by 49 study participants.}
\label{fig:goal_frequency}
\Description{Fully described in the text.}
\end{figure}

The survey results indicate that all goals extracted from the interview study are also pursued by the survey participants (see Figure~\ref{fig:goal_frequency}).
More than 95\% of the survey participants pursued the goal \emph{increase \ac{ML} model performance} and 75\% aimed to achieve \emph{decrease necessary computations}. 75\% sought to \emph{decrease practitioner effort}.
67\% of the practitioners aimed to \emph{increase \ac{ML} model understanding}.
The least pursued goals are \emph{satisfy requirements} (63\%) and \emph{comply with target audience} (53\%).

\begin{figure}[t]
\includegraphics[width=0.5\linewidth]{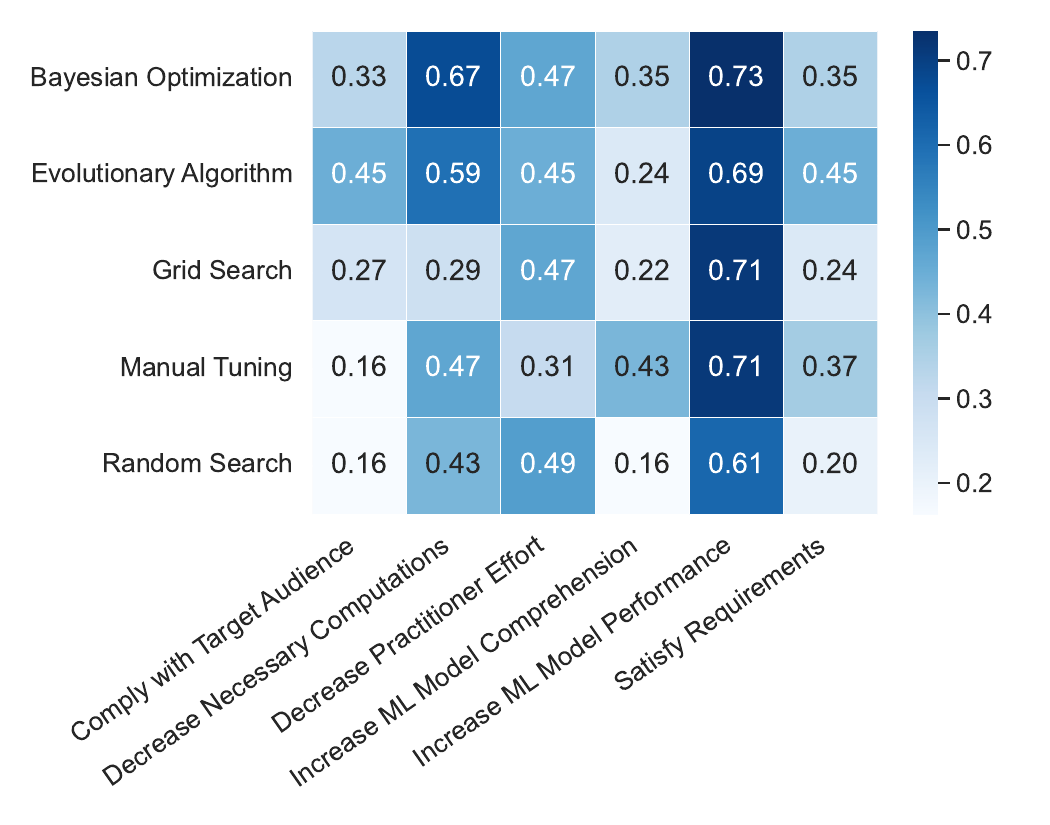}
\centering
\caption{Frequency of goal and \ac{HPO} method combinations. Per cell, all presented values are normalized to the number of participants having applied the corresponding \ac{HPO} method.}
\label{fig:method_goals}
\Description{Fully described in the text.}
\end{figure}

Figure~\ref{fig:method_goals} shows how often the survey participants used \ac{HPO} methods to reach the goals identified in the interview study.
67\% of the survey participants tried to decrease necessary computations by using Bayesian optimization and 59\% using evolutionary algorithms.
About 50\% of the participants tried to decrease the necessary computations by applying manual tuning.
Random search and grid search were least often used, with 43\% and 29\%, respectively.
Decreasing practitioner effort was of interest for less than 50\% of the participants, with very similar responses for all \ac{HPO} methods (yet a notable exception of manual tuning with only 31\%).
The survey participants primarily used manual tuning to enhance their comprehension of \ac{ML} models.
Interestingly, participants also tried to use Bayesian optimization twice as often as grid search and random search to achieve this goal despite its black-box nature.
Participants from academia have only tried to satisfy requirements or be compliant with target audience half as often as participants from industry.

\textit{Increasing \ac{ML} model performance} is the most common goal in \ac{HPO} that is pursued across all \ac{HPO} methods taken into account in this study. Most often, practitioners used Bayesian optimization to achieve this goals.

Practitioners mostly aimed to \textit{satisfy requirements} using evolutionary algorithms (45\%), Bayesian optimization (35\%), and manual tuning (37\%).
The study participants less often used grid search (24\%) and random search (20\%) to achieve this goal.

\emph{Comply with target audience} was the least pursued goal for the participants.
About 45\% of the participants used evolutionary algorithms for this goal.
We could not uncover differences between manual tuning and Bayesian optimization, with 30\% of the participants using the respective method.
Only random search and manual tuning were very seldom used to achieve this goal (about~15\%).

In contrast to their academic counterparts, industry participants rarely used Bayesian optimization or evolutionary algorithms to increase model understanding. Participants from academia preferred to use grid search to reduce their effort. Additionally, industry practitioners more frequently relied on manual tuning to meet specific requirements.

It is evident that participants employed different \ac{HPO} methods to pursue similar goals, making it difficult to establish a clear mapping between individual methods and goals. For example, while one interviewee selected manual tuning to \textit{increase \ac{ML} model understanding}, another favored Bayesian optimization to pursue the same goal.

\myquote{Interviewee~\#9, \ac{ML} Innovator in Academia}{Because especially when entering new areas, we would like to understand step by step what is working and what is not.}

\myquote{Interviewee~\#15, \ac{ML} Innovator in Industry}{I almost always select Bayesian optimization to get an idea in which region I find the [hyper-] parameter [values].}

The ambiguous responses suggest that the choice of \ac{HPO} methods is not simply explainable based on practitioners' goals. For example, practitioners used different \ac{HPO} methods to reach identical goals. This underscores the complexity of understanding why practitioners choose specific \ac{HPO} methods.
To better understand practitioners' reasons for selecting \ac{HPO} methods, contextual factors that influence practitioners in their decisions for \ac{HPO} methods to reach goals seem important.
By also illuminating contextual factors in combination with the use of \ac{HPO} methods and goals to be achieved, the motives of practitioners to use \ac{HPO} methods should be better understood.

\subsubsection{Contextual Factors That Influence HPO Method Selections}
\label{subsec:results-dec-fac}

We identified fourteen contextual factors that can influence practitioner decisions for using \ac{HPO} methods to achieve specific goals. The contextual factors can be grouped into three themes (see Table~\ref{tab:decision_factors}): \textit{own knowledge}, \textit{social environment}, and \textit{technical environment}.

\paragraph{Own Knowledge} Practitioner decisions for \ac{HPO} methods are influenced by practitioners' internal knowledge of \ac{HPO} and \ac{ML} models.
We identified three contextual factors related to own knowledge: \emph{\ac{HPO} method comprehension}, \emph{\ac{ML} model understanding}, and \emph{personal experiences}.

\begin{table*}[b]
    \renewcommand{\arraystretch}{1.25}
    \centering

    \caption{Principal contextual factors related to own knowledge that can influence practitioner decisions for \ac{HPO} methods.}
    \label{tab:decision_factors}
    
    \begin{tabularx}{\textwidth}{@{} llX @{}}
    \toprule
     Theme & Contextual Factor & Description\\

     \midrule
     Own Knowledge &
     \makecell[lt]{\ac{HPO} Method Comprehension} &
     The self-perceived level of knowledge a practitioner has about the inner workings of an \ac{HPO} method\\
     &
     \makecell[lt]{\ac{ML} Model Comprehension} &
     The self-perceived degree of understanding of the inner workings of an \ac{ML} model with which a practitioner explains changes in the behavior of the \ac{ML} model caused by altering hyperparameter values\\
     &
     \makecell[lt]{Personal Experience} &
     The available internal knowledge of a practitioner that has been generated by past activities (\eg~personal best practices to solve a specific problem type)\\

     \midrule
     
     Social Environment &
     \makecell[lt]{Acceptance of Advanced  \\\ac{HPO} Methods} &
     The extent to which advanced \ac{HPO} methods (\eg~Bayesian optimization) are valued by a target group\\
     &
     Literature &
     The knowledge acquired on the basis of published text documents (\eg~articles, blog entries, and papers)\\
     &
     Shared Opinions &
     The knowledge acquired on the basis of advice from peers (\eg colleagues)\\
     &
     \makecell[lt]{Tension for Resources} &
     The degree to which constrained compute resources cause conflicts between practitioners regarding the allocation of those resources\\

     \midrule     
     Technical Environment &
     \makecell[lt]{Available Compute Resources} &
     The amount of compute resources available for \ac{HPO}\\    
     &
     \makecell[lt]{Cost of Objective Function} &
     The amount of compute resources required to evaluate a single point within a hyperparameter value space\\    
     &
     \makecell[lt]{\ac{HPO} Method Traceability } &
     The extent to which a sequence of sample points can be backtraced or predicted\\    
     &
     \makecell[lt]{\ac{HPO} Setup Readiness } &
     The degree to which an \ac{HPO} tool and associated test environments are ready to use (\eg preinstalled \ac{HPO} tools on a cluster)\\    
     &
     \makecell[lt]{Parallelization Possibilities} &
     The degree to which multiple independent \ac{ML} models can be simultaneously evaluated\\    
     &
     \makecell[lt]{Search Space Size} &
     The number of possible hyperparameter configurations\\    
     &
     \makecell[lt]{Usability of \ac{HPO} Tools } &
     The perceived ease with which practitioners achieve their goals by using an \ac{HPO} method and corresponding tools\\
     
    \bottomrule
    \end{tabularx}
\end{table*}

\emph{\ac{HPO} method comprehension} refers to the degree to which practitioners understand how \ac{HPO} methods work and how to apply them.
Practitioners tend to neglect \ac{HPO} methods they do not sufficiently understand.
For example, two interviewees stated they had disregarded Bayesian optimization because they felt they did not sufficiently understand its inner workings.
One interviewee from the industry perceived grid search to be faster to implement and easier to use compared to Bayesian optimization because using the latter would have required the interviewee to learn an \ac{HPO} method they were not experienced with.
Another interviewee perceived random search as uncontrolled, which caused them to decide against it.
Two interviewees decided to use grid search because they perceived grid search as easy to understand and implement.

\emph{\ac{ML} model comprehension} refers to a practitioner's ability to explain changes in an \ac{ML} model's behavior caused by altering hyperparameter values based on an understanding of the inner workings of the \ac{ML} model.
The perceived degree of \ac{ML} model understanding plays an important role.
Interviewees who perceived their \ac{ML} model understanding as high stated to have chosen manual tuning.
Due to their thorough \ac{ML} model understanding, those interviewees claimed that they can find appropriate hyperparameter configurations without extensive \ac{HPO}.
The interviewees perceived programmatic \ac{HPO} methods as not taking advantage of known effects of hyperparameters on \ac{ML} model development and performance. 
Examples of such known effects include a high learning rate, which can accelerate learning but may lead to instability; a low learning rate, which slows convergence; strong regularization, which can cause underfitting; and weak regularization, which can lead to overfitting \cite{yang2020hpo}.

\myquote{Interviewee~\#1, \ac{ML} Innovator in Academia}{Effects of hyperparameters are often deducible, but optimizers [here: \ac{HPO} methods] usually do not support functionalities for this.}

Interviewees who deemed their \ac{ML} model understanding as low tended to use random search or Bayesian optimization.
Low \ac{ML} model understanding made it difficult for interviewees to predict the challenges they would encounter in \ac{HPO}.
To better react to unforeseen challenges, interviewees stated to rather choose manual tuning.
For example, manual tuning can facilitate spotting and correcting mistakes when errors occur during development of novel \ac{ML} models type because feedback loops are faster compared to those of programmatic \ac{HPO} methods:

\newpage

\myquote{Interviewee~\#3, \ac{ML} Engineer in Academia}{Because we altered the standard architecture as a whole, we were not really sure what problems we would face. So that was one of the reasons to stick with manual tuning.}

\emph{Personal experiences} refers to the available internal knowledge that a practitioner generated through past activities (\eg~personal best practices for solving a specific type of problem).
The interviewees stated to tend to use \ac{HPO} methods with which they had positive experiences:

\myquote{Interviewee~\#2, \ac{ML} Innovator in Academia}{I have also had good experiences with it [here: Bayesian optimization] in a previous paper }

\paragraph{Social Environment} Choices for \ac{HPO} methods are influenced by the social environment of practitioners, especially by four contextual factors: \emph{acceptance of advanced \ac{HPO} methods}, \emph{literature}, \emph{shared opinions}, and \emph{tension for resources}.

\emph{Acceptance of advanced \ac{HPO} methods} refers to the extent to which advanced \ac{HPO} methods, such as Bayesian optimization, are valued by a target group.
Low acceptance of advanced \ac{HPO} methods in a community targeted by a practitioner can make them avoid extensive \ac{HPO} entirely and choose manual tuning.
For example, an academic stated that they perceived the use of advanced \ac{HPO} methods and extensive \ac{HPO} as not being valued by their community.
According to the interviewee, their community encourages the use of pre-trained \ac{ML} models in combination with manual fine-tuning to avoid extensive \ac{HPO}.
Although the interviewee perceived Bayesian optimization as more suitable for increasing \ac{ML} model performance, they felt discouraged by the attitude of their community and applied manual tuning instead.


\emph{Shared opinions} cover external knowledge acquired on the basis of advice from peers (\eg~colleagues).
The interviewees explained to have chosen \ac{HPO} methods that are considered as commonly used in their labs or by their peers.
In various communities, different \ac{HPO} methods are applied so frequently that their use becomes habitual.
For example, manual tuning was commonly used in one research group, while Bayesian optimization was considered the primarily applied \ac{HPO} method in another one.
The interviewees associated with those communities applied the respectively manifested \ac{HPO} methods.
This indicates that the immediate social environment has a noticeable influence on practitioners' \ac{HPO} method choices.

\emph{Literature} refers to external knowledge acquired on the basis of published text documents (\eg~articles, blog entries, papers).
Practitioners, from academia and industry alike, are guided in their choices of \ac{HPO} methods by recommendations from literature on \ac{ML} models similar to their own.
All practitioners, who primarily based their decisions on literature, chose Bayesian optimization that attests high sample efficiency (\eg \cite{turner_bayesian_2021}).

\emph{Tension for shared resources} refers to the degree to which constrained compute resources cause conflicts between practitioners.
Availability of only shared resources can cause tensions among colleagues, for example, when practitioners must compete for compute resources to perform \ac{HPO}.
Such tensions led one academic scientist to choose manual tuning to avoid arguing with colleagues over compute resources.

\paragraph{Technical Environment}

Contextual factors associated with the technical environment refer to technical constraints (\eg caused by insufficient compute resources) that influence practitioners' selections of \ac{HPO} methods.
The interviewees stated seven contextual factors associated with the technical environment: \emph{available compute resources}, \emph{cost of objective function}, \emph{\ac{HPO} method traceability}, \emph{\ac{HPO} setup readiness}, \emph{parallelization possibilities}, \emph{search space size}, and \emph{usability of \ac{HPO} tools}.

\emph{Available compute resources} refers to the amount of compute resources available for \ac{HPO}. Practitioners choose manual tuning when faced with constrained available compute resources. They perceive that in combination with a high degree of \ac{ML} model understanding, they can outperform programmatic \ac{HPO} methods.
If the available compute resources are too scarce, exploration of large search spaces is hard.
Practitioners need to reduce search spaces, decrease the number of necessary function evaluations, or decrease computational cost per function evaluation (\eg~by low-fidelity approximations) to be able to perform \ac{HPO}.
To decrease the number of necessary function evaluations, three scientists stated to have used manual tuning because they were able to predict the influence of hyperparameter values on \ac{ML} model performance.
Given constrained available compute resources and a sufficient degree of \ac{ML} model understanding, scientists in this study perceived manual tuning as superior to Bayesian optimization and random search.

Two interviewees chose \ac{HPO} methods depending on the \emph{cost of the objective function} they sought to optimize (\ie~training of an \ac{ML} model).
The cost of the objective function refers to the amount of compute resources required to evaluate a single point within the hyperparameter value space.
Similar to constrained compute resources, the interviewees chose manual tuning when faced with too expensive objective functions.
When the interviewees perceived their degree of \ac{ML} model understanding as high, they deemed manual tuning more efficient.

\emph{\ac{HPO} method traceability} refers to the extent to which a sequence of sample points can be backtracked or predicted by the practitioner, which requires that the selection of samples by the \ac{HPO} method is comprehensible for and reproducible by practitioners.

\emph{\ac{HPO} setup readiness} refers to the degree to which \ac{HPO} tools and test environments are ready to use (\eg~preinstalled \ac{HPO} tools on the cluster). Some of the study participants stated to be unwilling to set up new \ac{HPO} tools but rather use already set up tooling, regardless of the quality produced by the corresponding \ac{HPO} method.

\emph{Parallelization possibilities} of \ac{HPO} methods refer to the degree to which independent \ac{ML} models can be simultaneously evaluated. Parallelization possibilities can be constrained by, for example, too few software licenses.
Two interviewees chose Bayesian optimization if parallelization of \ac{HPO} was not possible due to a misconception of the sequential proceeding in Bayesian optimization.
Another interviewee stated that they chose Bayesian optimization if their objective function is expensive and \ac{HPO} parallelization is not possible.

\emph{Search space size} refers to the number of possible hyperparameter combinations.
The interviewees stated that the number of hyperparameters included in the \ac{HPO} impacted their decisions for \ac{HPO} methods. For example, they stated to opt for random search over grid search and Bayesian optimization if the number of hyperparameters is large.

\emph{Usability of \ac{HPO} tools} refers to the perceived ease with which practitioners can achieve their goal by using an \ac{HPO} method and corresponding implementations. Interviewees from industry remarked that many advanced \ac{HPO} tools are probably useful for academic purposes but lack the necessary level of maturity to be viable in practice. Within the scope of usability, practitioners demanded more automation of cumbersome tasks in \ac{HPO} such as infrastructure orchestration:

\myquote{Interviewee~\#14, \ac{ML} Innovator in Industry}{What beats everything for me is that I have a dashboard that's somewhere in the cloud that orchestrates my various agents, where I can sort of say online, `Start another agent on this machine,' or that on the machine I just have to say, `Start another agent on this sweep here,' and I don't have to worry about the agents talking to each other or having a shared database running on some cluster. This functionality, it overrides everything. If I had some mega highly optimized Bayesian optimization tool that didn't have that functionality, I wouldn't use it.}

\begin{figure}[t]
\includegraphics[width=0.8\linewidth]{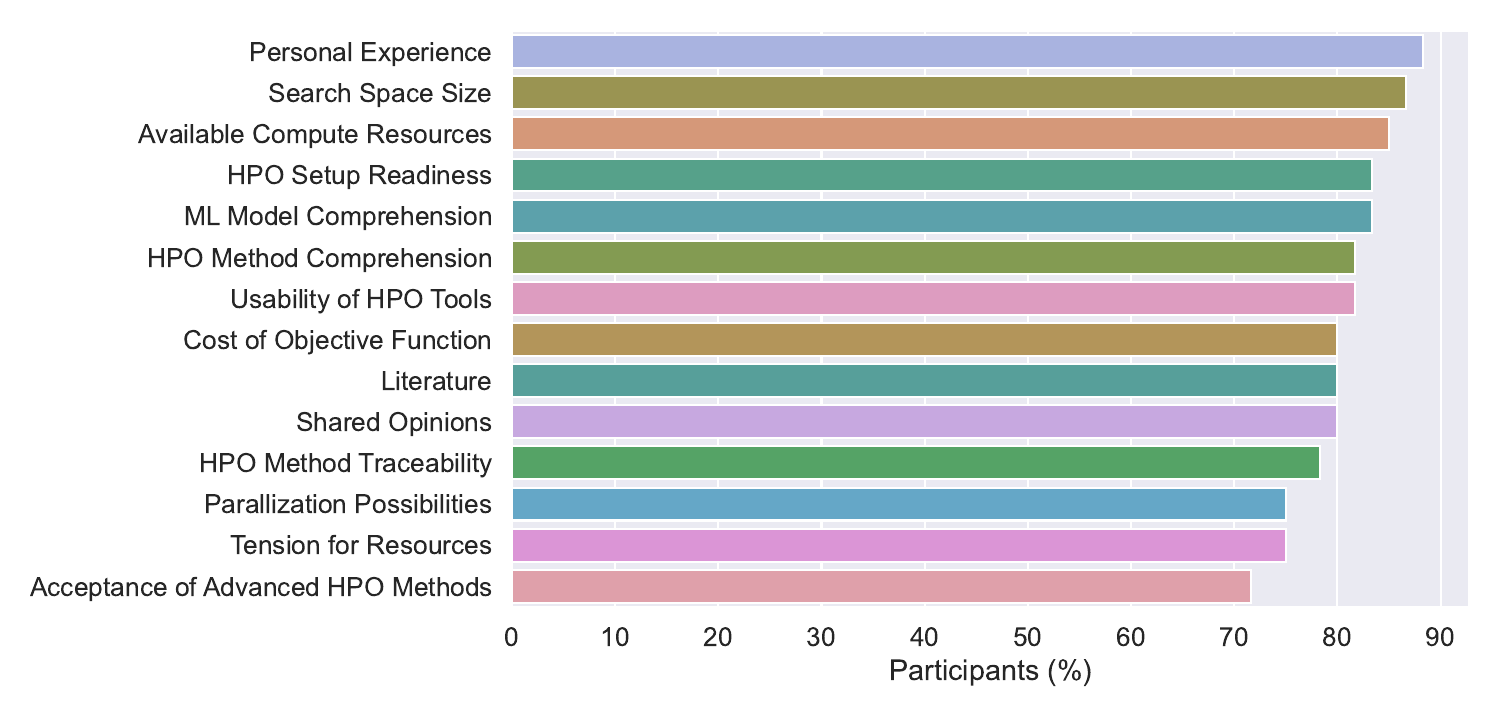}
\centering
\caption{Percentage of participants that incorporated the individual contextual factors.}
\label{fig:used_context_factors}
\Description{Fully described in the text.}
\end{figure}

The results of the survey study show that each contextual factor influenced at least 70\% of the participants in selecting \ac{HPO} methods (see Figure~\ref{fig:used_context_factors}).
More than 85\% of the survey participants considered the decision factors \emph{personal experience}, \emph{search space size}, and \emph{available compute resources} in their selections of \ac{HPO} methods.
The contextual factors least considered are \emph{acceptance of advanced methods}, \emph{tension for resources}, and \emph{parallelization possibilities}. They were only relevant for less than 75\% of the survey participants in their past \ac{ML} projects.
The remaining contextual factors from all three themes, including \emph{shared opinions}, \emph{model comprehension}, and \emph{\ac{HPO} method traceability}, have been considered by 75--85\% of the survey participants.

\begin{figure}[h]
\includegraphics[width=0.8\linewidth]{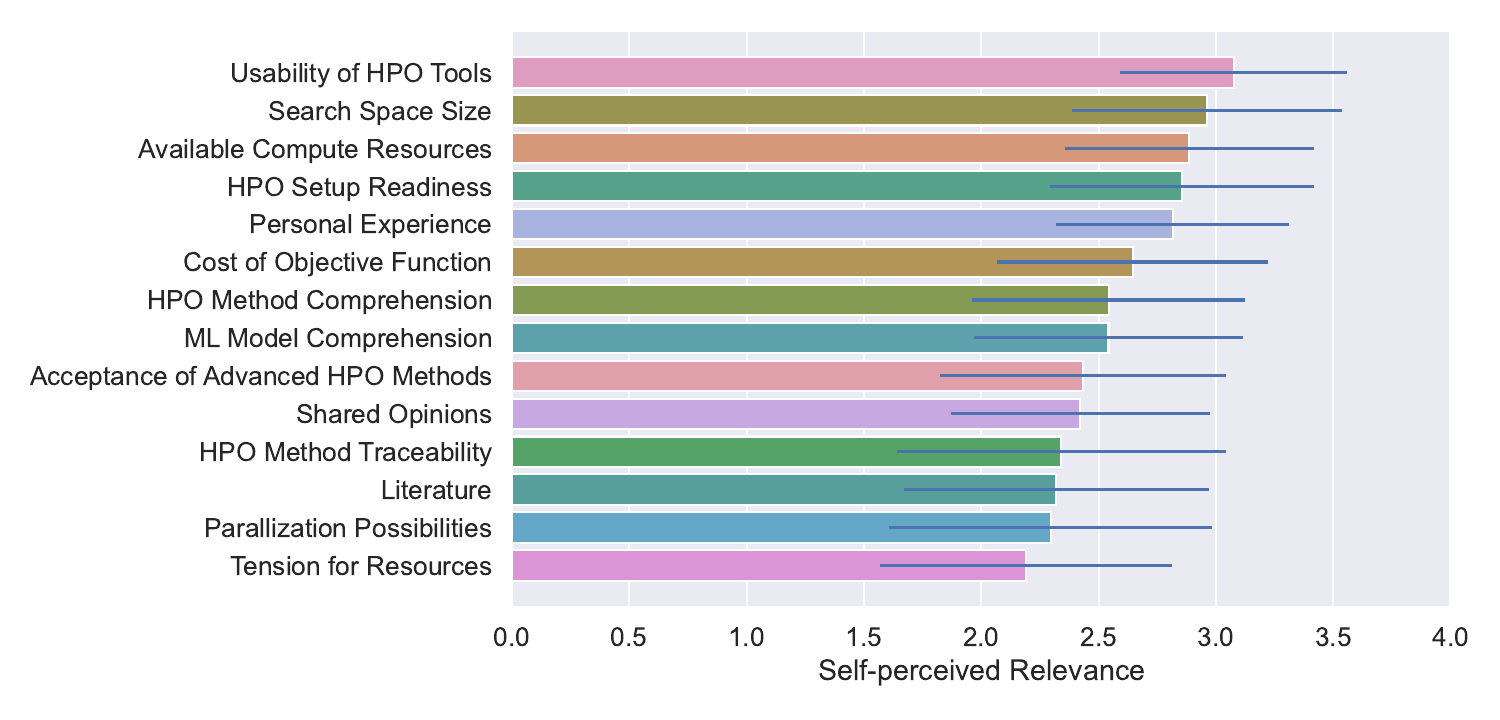}
\centering
\caption{Overview of the average self-perceived relevance of contextual factors. Results are reported on a scale from 0 (very low) to 5 (very high). On average, no contextual factor was rated with a higher relevance than 3, which is why we masked out the values 4 and 5. Blue lines indicate error bars of one standard deviation.}
\label{fig:context_factor_importance}
\Description{Fully described in the text.}
\end{figure}

The identified contextual factors are of different self-perceived relevance for the selection of \ac{HPO} methods (see Figure~\ref{fig:context_factor_importance}).
The self-perceived relevance of contextual factors interdepends with the consideration of contextual factors.
\emph{Usability of \ac{HPO} tools} and \emph{search space size} are the most relevant contextual factors, closely followed by \emph{available compute resources} and \emph{\ac{HPO} setup readiness}.
Other relevant contextual factors are \emph{personal experience}, \emph{cost of the objective function}, and the \emph{\ac{HPO} method} and \emph{\ac{ML} model comprehension}.
All contextual factors associated with the social environment are less relevant for the survey participants, with \emph{tension for resources} being considered the least.

\begin{figure*}[h]
    \centering
    \subfloat[Bayesian optimization]{\includegraphics[width=0.48\textwidth]{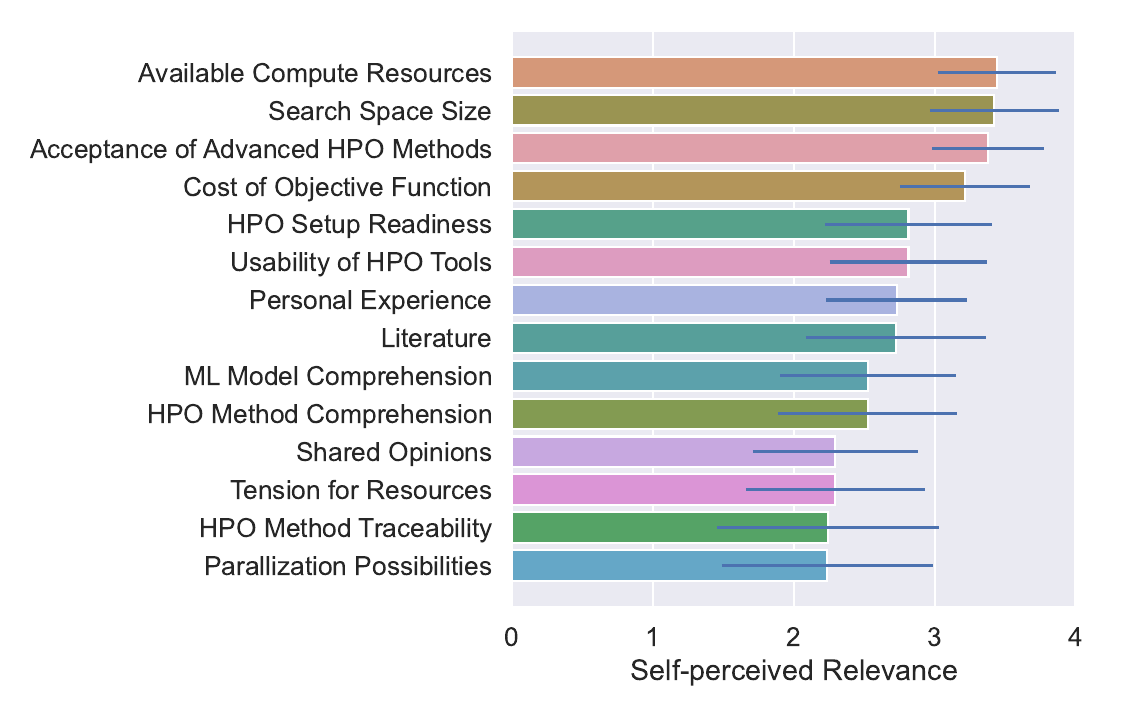}%
    \label{fig:bayesian-optimization}}
    \hfill
    \subfloat[Grid search]{\includegraphics[width=0.48\textwidth]{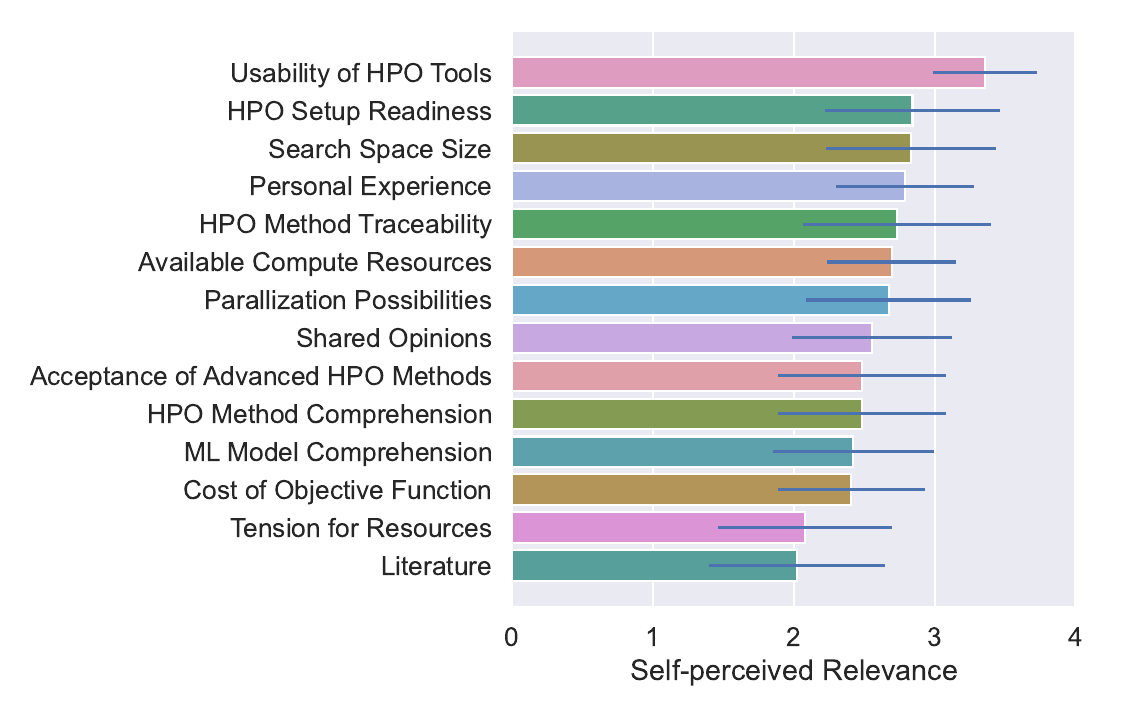}%
    \label{fig:grid-search}}
    \hfill
    \subfloat[Manual tuning]{\includegraphics[width=0.48\textwidth]{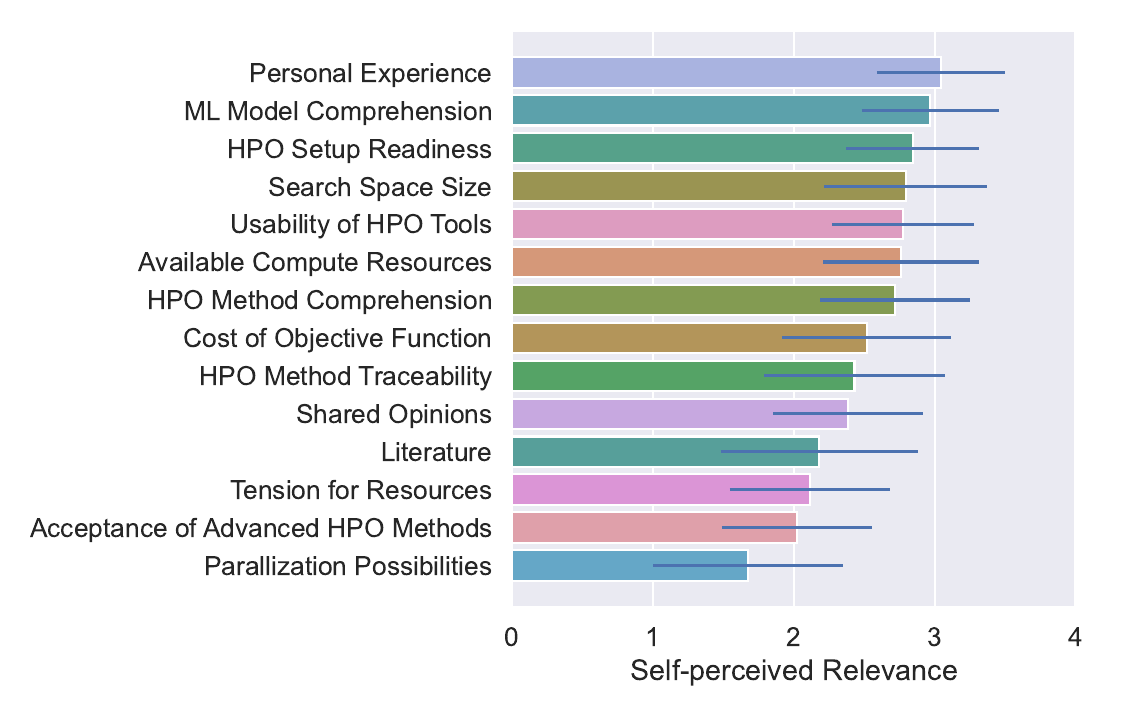}%
    \label{fig:manual-tuning}}
    \hfill
    \subfloat[Random search]{\includegraphics[width=0.48\textwidth]{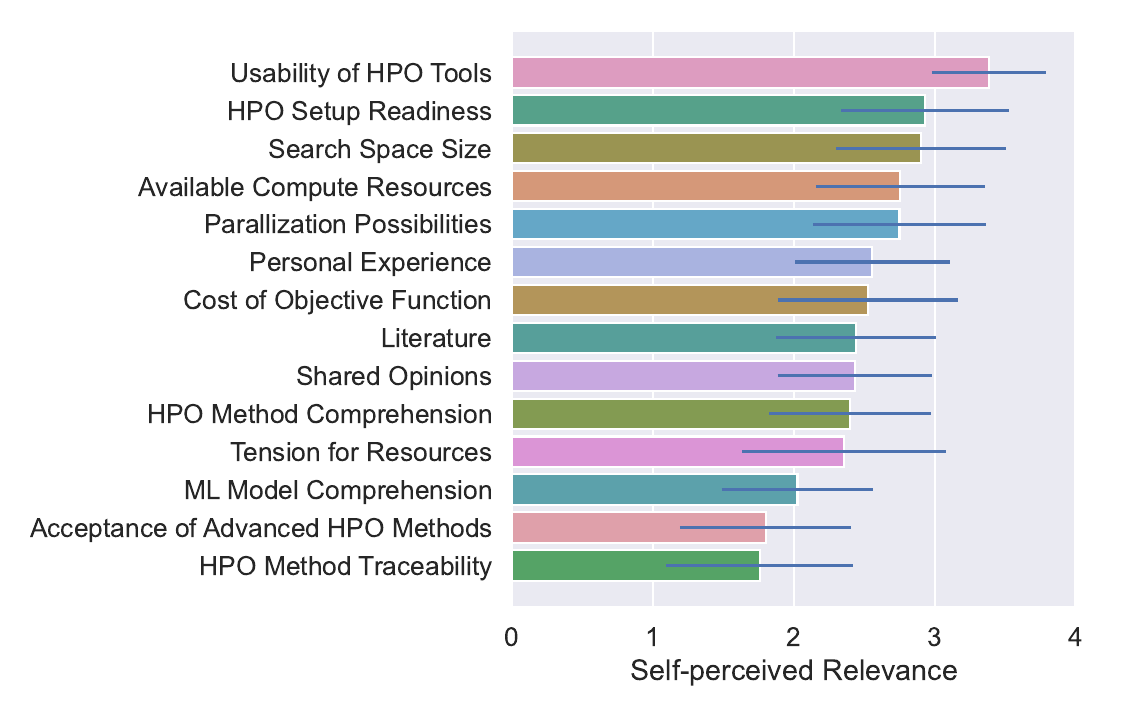}%
    \label{fig:random-search}}
    
    \caption{Overview of the average self-perceived relevance of contextual factors per \ac{HPO} method. Results are reported on a scale from 0 (very low) to 5 (very high). Blue lines indicate error bars of one standard deviation.}
    \label{fig:context_factors_per_HPO_method}
    \Description{Fully described in the text.}
\end{figure*}

Figure~\ref{fig:context_factors_per_HPO_method} illustrates the self-perceived relevance of contextual factors for each of the \ac{HPO} methods in the scope of this study.
The survey participants considered the \emph{available compute resources}, \emph{search space size}, \emph{acceptance of advanced methods}, and \emph{cost of the objective function} mostly when selecting Bayesian optimization.
This aligns with usage reasons of Bayesian optimization commonly proclaimed in literature~\cite{Hutter2018b, souza2021bayesian, garnett_bayesoptbook_2022}.
The selection of grid search was mostly influenced by \emph{usability of \ac{HPO} tools}, \emph{\ac{HPO} setup readiness}, and \emph{search space size} with similar results for random search.
A potential explanation could be the availability of these \ac{HPO} methods in established and publicly available \ac{ML} libraries like \textit{scikit-learn}.
Moreover, survey participants considered their \emph{personal experience}, \emph{\ac{ML} model comprehension}, and \emph{\ac{HPO} setup readiness} most relevant when selecting manual tuning, making \emph{own knowledge} more important than \emph{technical environment}.
This indicates that the relative importance of contextual factors in a specific instance leads to different selections of \ac{HPO} methods.

\subsection{Perceived Success of Using HPO Methods to Achieve Specific Goals}
\label{subsec:results-perceived-success}

Practitioners seem to have diverse and individual motivations for using specific \ac{HPO} methods.
However, these choices do not always lead to the desired outcomes. To distinguish between successful and unsuccessful experiences of practitioners in using \ac{HPO} to reach their goals, we asked the study participants to what extent they perceive to have attained which goals using what \ac{HPO} methods.

\begin{figure}[h]
    \includegraphics[width=0.6\linewidth]{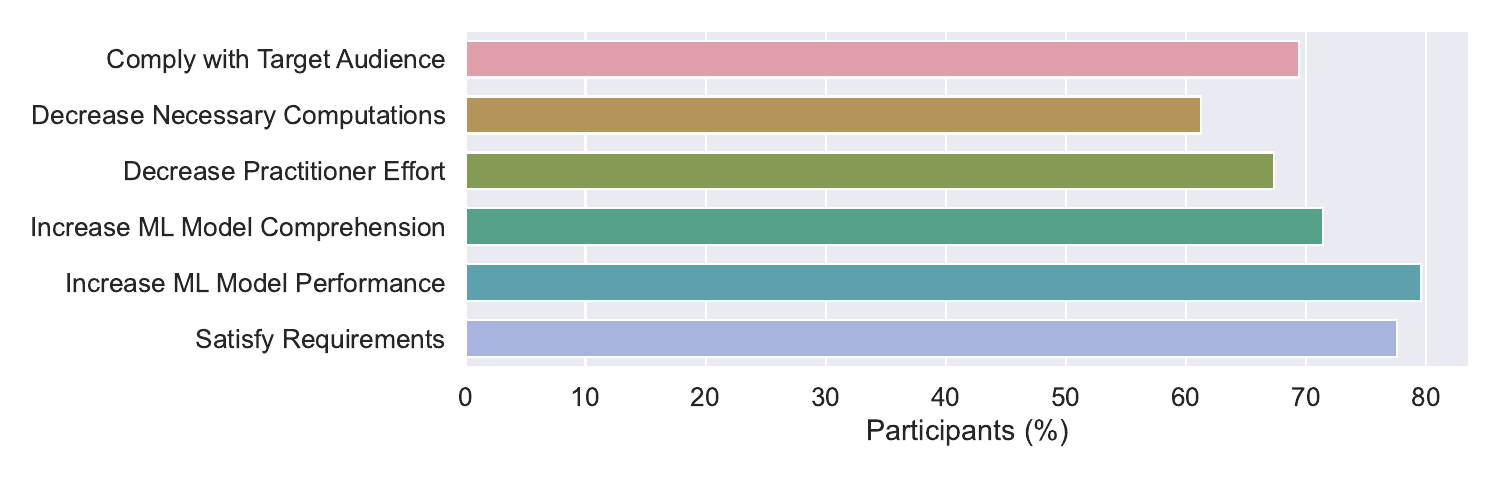}
    \centering
    \caption{Self-reported success rate per goal by 49 study participants.}
    \label{fig:success_rate_per_goal}
    \Description{Fully described in the text.}
\end{figure}

Figure~\ref{fig:success_rate_per_goal} shows the success rate per goal perceived by the survey participants.
Roughly 75\% of the participants responded to have successfully \emph{increased \ac{ML} model performance}, reached \textit{complied with target audience}, \emph{increase \ac{ML} model understanding}, or \emph{satisfied requirements}.
Industry participants were successful in trying to \emph{satisfy requirements} or \emph{comply with target audience}.
67\% of the participants stated that they were able to achieve the goal \emph{decrease practitioners effort}.
Only 62\% of participants considered themselves successful in \textit{reducing computational requirements}.

\begin{figure}[b]
    \includegraphics[width=0.5\linewidth]{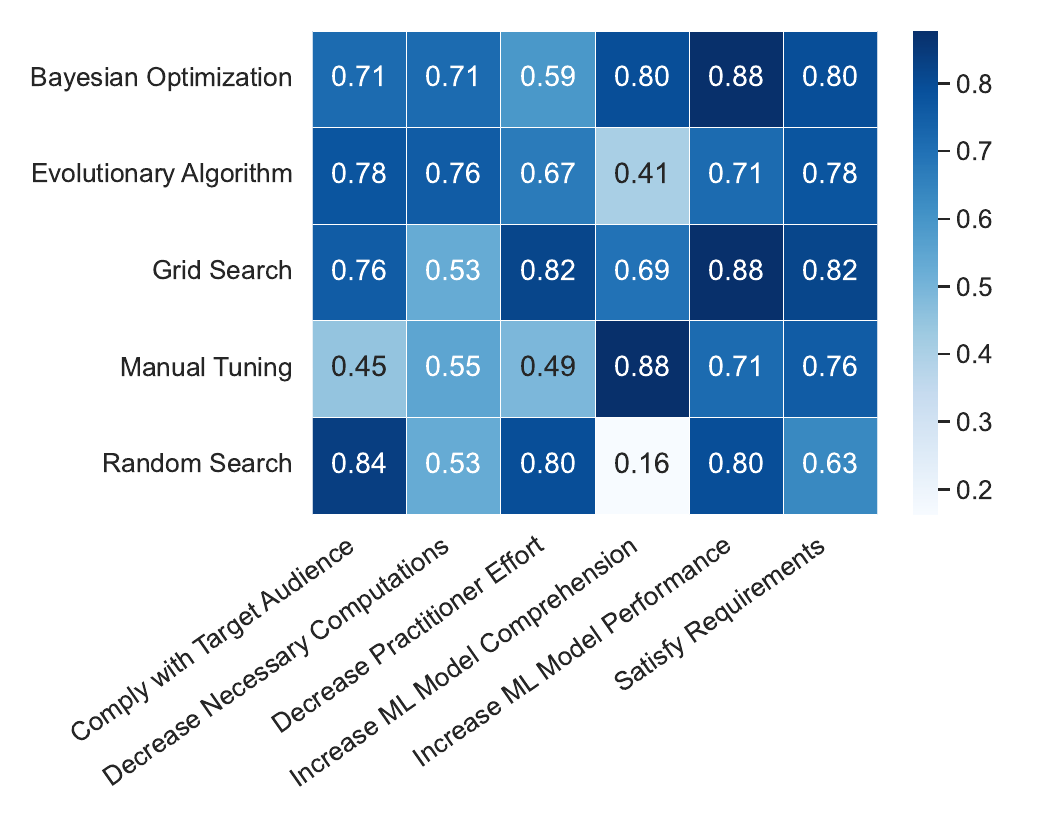}
    \centering
    \caption{Self-reported success rates per goal-method combination by 49 study participants.}
    \label{fig:success_rate}
    \Description{Fully described in the text.}
\end{figure}

The self-perceived success rates strongly vary between combinations of goals and \ac{HPO} methods (see Figure~\ref{fig:success_rate}).
Survey participants reported lower success rates in \emph{reducing computational requirements} when using manual tuning, grid search, or random search.
They perceived themselves as rather successful in reaching this goal when using Bayesian optimization or evolutionary algorithms.
\emph{Decreasing practitioner effort} was best achieved using grid search or random search according to the survey participants.
Bayesian optimization and evolutionary algorithms were perceived as less effective in decreasing effort.
A potential explanation could be that those \ac{HPO} methods often require more effort to be set up compared to others \cite{Khuat2022}.
Moreover, participants also perceived manual tuning as ineffective in decreasing their efforts.
Participants perceived manual tuning as very helpful to \emph{increase \ac{ML} model understanding.}
Even though Bayesian optimization is considered a black-box optimization technique~\cite{Frazier2018}, it was also perceived as suitable to increase \ac{ML} model understanding.
Random search, grid search, and evolutionary algorithms were perceived as unsuitable for increasing \ac{ML} model understanding.
Most participants did not perceive noteworthy differences between the effectiveness of \ac{HPO} methods to successfully \textit{increase \ac{ML} model performance} and to \emph{satisfy requirements}.
Only evolutionary algorithms were perceived as significantly more successful in trying to satisfy requirements.
Survey participants successfully used grid search, random search, evolutionary algorithms, and Bayesian optimization to \emph{meet the expectations of their target audience}.
Manual tuning was applied with lower success rates for this goal.

\section{Discussion}
\label{sec:discussion}
The results presented in the previous section offer novel insights into how practitioners use \ac{HPO} methods. In this section, we highlight and discuss the principal findings from those insights. We describe how the results of this study contribute to more user-centric development of \ac{HPO} methods and tools, explicate the limitations of this study, and outline future research directions.

\subsection{Principal Findings}

Besides improving \ac{ML} model performance in the first place, practitioners are second most interested in decreasing the required amount of necessary computations and their personal efforts to perform \ac{HPO}, which reflects the basic motivation for development of \ac{HPO} tools \cite{Wang2021}.
To decrease necessary computations, the study participants predominantly used Bayesian optimization. The frequent use of Bayesian optimization to decrease necessary computations suggests that practitioner perceptions of the benefits of Bayesian optimization are coherent with its benefits empirically shown in prior research~\cite{turner_bayesian_2021}.

Notwithstanding the high sample efficiency of Bayesian optimization, some practitioners prefer to use manual tuning to decrease the number of necessary computations.
In particular, if practitioners assume that their \ac{ML} model understanding is high, they expect to outperform Bayesian optimization.
Yet, it is difficult to compare manual tuning to programmatic \ac{HPO} methods due to its reliance on a mixture of explicit and implicit knowledge that often cannot be fully extracted from observations of practitioner actions. One way to leverage \ac{ML} model understanding could be to integrate practitioner priors on the location of well-performing hyperparameter configurations into Bayesian Optimization effectively warm-starting optimizations~\cite{souza2021bayesian, Hvarfner2022, MallikBHSJLNH23}.

In addition to well-established goals pursued in \ac{HPO}, this study presents a multitude of goals pursued by practitioners that are less emphasized in research on \ac{AutoML}. Practitioners have strong motives for \ac{HPO} beyond improving \ac{ML} model performance.
For example, the results of this study show that various practitioners are interested in better understanding their subject of work, including \ac{HPO} methods and \ac{ML} models, prior to using them.
To better understand \ac{ML} models, most practitioners chose \ac{HPO} methods they understood over methods they would have to study first. This led practitioners to opt for manual tuning instead of programmatic \ac{HPO} methods.
Many participants perceived programmatic \ac{HPO} tools as unsuited to increase \ac{ML} model understanding.
Although numerous software packages for advanced \ac{HPO} methods are available for out-of-the-box use without requiring an understanding of their internal workings (\eg SMAC3 \cite{Lindauer2022} or Optuna \cite{AkibaSYOK19}), practitioners appear reluctant to adopt \ac{HPO} methods they do not fully understand.
Practitioners tend to rely on their own knowledge rather than giving up control to insufficiently understood \ac{ML} models and \ac{HPO} tools.

To help practitioners increase \ac{ML} model understanding, various software tools were designed.
Such tools, predominantly \ac{HPO} tools in \ac{AutoML}, mainly focus on supporting measurements of influences of hyperparameter configurations on \ac{ML} model performance (\eg \cite{Hutter2014, biedenkapp-aaai17, Moosbauer2021,segel-automl23a}).
With a focus on \ac{HPO} methods, especially tools for visual analytics are envisioned to support better understanding of internal behaviors of \ac{HPO} methods by visualizations (\eg \cite{Golovin2017, Biedenkapp2019,sass-realm22a, Zoller2022}).
Despite the existence of such tools, practitioners tend to prefer to use manual tuning, which may have different reasons. The first reason may be that practitioners are unaware of \ac{HPO} tools that can help increase \ac{ML} model understanding.
A second reason may be that \ac{HPO} tools do not fulfill the information needs of practitioners to increase \ac{ML} model understanding because \ac{HPO} tools mainly focus on performance of \ac{ML} models, which is, as shown in this study, only one practitioner motive in \ac{HPO}.
A third reason may be that the functioning of \ac{HPO} tools themselves is hard to comprehend for practitioners (\eg~because such \ac{HPO} tools implement unfamiliar and complex \ac{HPO} methods), which leads practitioners to prefer \ac{HPO} methods they are familiar with.

\begin{figure*}[ht]
    \centering
    \subfloat[Academia (27 participants).]{\includegraphics[width=0.49\textwidth]{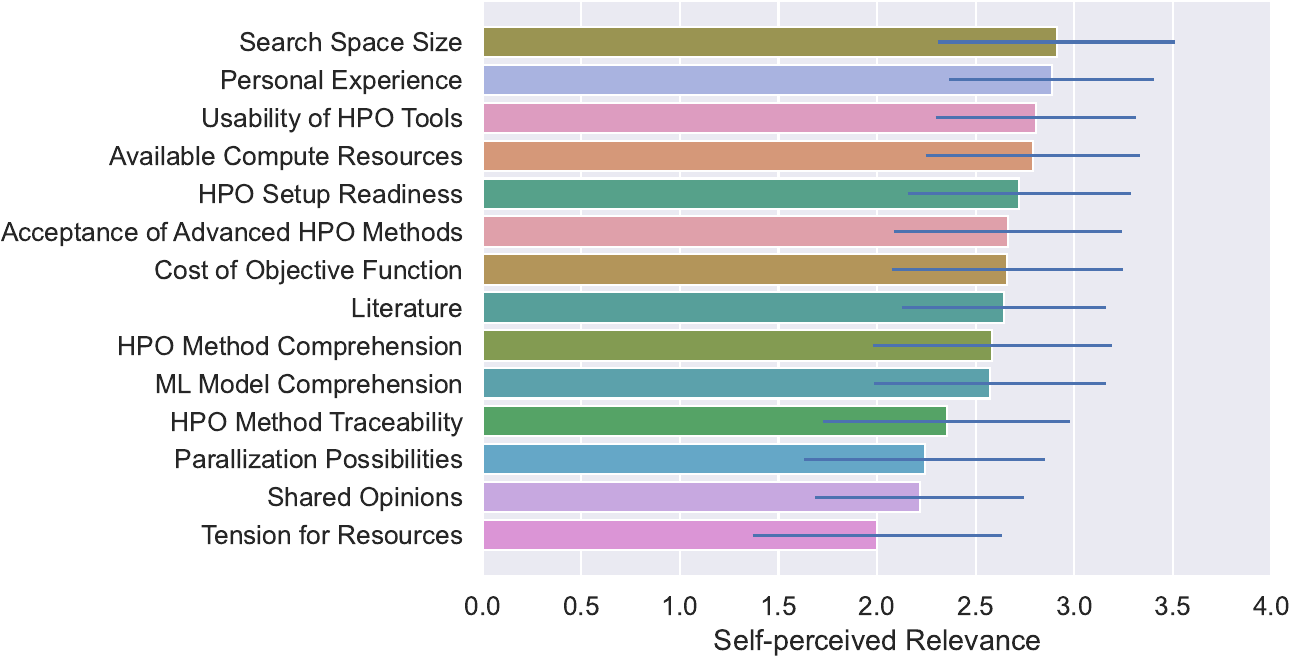}%
    \label{fig:relevance_contextual_factors_academia}}
    \hfill
    \subfloat[Industry (22 participants).]{\includegraphics[width=0.49\textwidth]{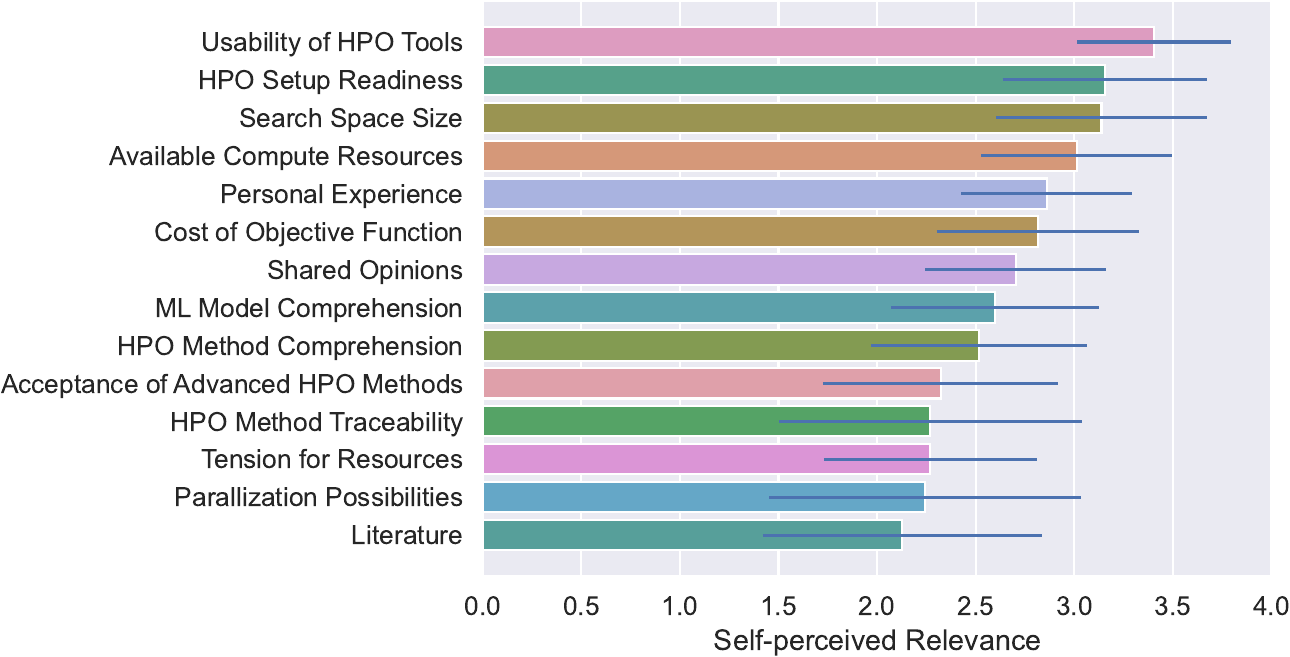}}
    
    \caption{Relevance of the identified contextual factors in academia and industry.}
    \label{fig:ovw-relevance-contextual-factors}
    \Description{Fully described in the text.}
\end{figure*}

The results presented in this study provide an aggregated view of practitioner motives in using \ac{HPO} methods. Although our analysis did not reveal distinct personas with significantly different usage patterns, our results highlight nuanced variations in how practitioners engage with \ac{HPO} methods. Specifically, we observed that the importance of goals and the use of \ac{HPO} methods remained relatively consistent among practitioners, regardless of their experience, age, or education level.
However, our findings indicate that contextual factors play a more prominent role in shaping the approaches of practitioners from academia versus those from industry (see Figure~\ref{fig:ovw-relevance-contextual-factors}).
Contrary to practitioners from industry, for example, academics tend to avoid using \ac{HPO} methods that are difficult to integrate into workflows or require prior training. This especially applies when using complex \ac{HPO} methods, according to the study participants.

To make the achievements of techno-centric \ac{HPO} research more actionable in practice and to support development of novel and more user-centered \ac{HPO} tools, it is essential to integrate practitioners' motivations for \ac{HPO} into the tool development process.
In response to calls for more human-centered \ac{AutoML} \cite{lindauer2024humancentered, lee2020human, sun2023automl, xin2021wither}, we propose improvements to programmatic \ac{HPO} methods based on our findings to complement technological advances in \ac{AutoML}, which primarily focus on conventional performance metrics, by incorporating social aspects.


\paragraph{Increase \ac{ML} Model Understanding}
The inspection of frequent \ac{HPO} method and goal combinations revealed that increasing \ac{ML} model comprehension is the only goal with a strongly higher association with manual tuning than programmatic \ac{HPO} methods (see Figure~\ref{fig:method_goals}).
A potential explanation could be that established \ac{HPO} tools usually focus on finding the best performing hyperparameters and do not provide details of other interesting hyperparameter values tested \cite{xin2021wither, Zoller2022}.
To overcome this limitation, \ac{HPO} tools should generate reports on the behavior of different \ac{ML} models, for example, the importance of individual hyperparameters.
For the generation of such reports, many methods are already available, such as functional ANOVA~\cite{Hutter2014}, ablation~\cite{biedenkapp-aaai17}, importance of local parameters~\cite{Biedenkapp2019}, partial dependence plots~\cite{Moosbauer2021}, and symbolic regressions~\cite{segel-automl23a}.
Alternatively, \ac{HPO} tools could provide additional insights about \ac{ML} model behavior.
Especially for more complex search spaces used for building complete \ac{ML} pipelines, information about the transformation of input data can additionally help increase \ac{ML} model understanding \cite{Zoller2022}.
Including such reports in \ac{HPO} tools can facilitate leveraging the benefits of advanced \ac{HPO} methods (\eg~high sample efficiency) while still helping practitioners to \emph{increase \ac{ML} model understanding}.

\paragraph{Explain \ac{HPO} Method Internals}
The results of this study show that a key barrier to the adoption of advanced \ac{HPO} methods is the perceived lack of transparency (in terms of \ac{HPO} method traceability) and interpretability (in terms of \ac{HPO} method comprehension), which affects the use of programmatic \ac{HPO} tools (see Figure~\ref{fig:bayesian-optimization}).
\ac{HPO} tools should provide more support explaining their internal behavior to make them better comprehensible to practitioners.
An easy approach would be simple visualizations of the hyperparameter values that were evaluated using parallel coordinates plots~\cite{Golovin2017}.
More sophisticated approaches could present information about the internals of their optimizers, for example, the surrogate model in Bayesian optimization~\cite{Biedenkapp2019}.
Such measures could help educate practitioners about \ac{HPO} methods and increase practitioners' confidence in the functioning and merits of programmatic \ac{HPO} tools.

\paragraph{Integrate Practitioners' \ac{ML} Model Understanding in HPO}
Figure~\ref{fig:manual-tuning} shows that \ac{ML} model comprehension is an important contextual factor for preferring manual tuning instead of a programmatic \ac{HPO} method.
Similarly, multiple interviewees mentioned that they prefer manual tuning when they are confident in predicting the impact of hyperparameters on the \ac{ML} model behavior.
To harness this knowledge, \ac{HPO} tools should enable practitioners to incorporate their relevant knowledge of behaviors of \ac{ML} models into \ac{HPO} tools prior to \ac{HPO} on a case-by-case basis, tailored to goals like \textit{improve \ac{ML} model performance}.
For example, practitioners could specify their perceived hyperparameter importance and influences between hyperparameters.
Furthermore, \ac{ML} models comprehension of practitioner could be directly incorporated into the search strategy of \ac{HPO} methods, for example to reduce search spaces \cite{xin2021wither} and better use available computing resources (the most important contextual factor for Bayesian optimization in Figure~\ref{fig:bayesian-optimization}).
Promising work in this direction includes methods for integrating prior knowledge into Bayesian optimization. This can be achieved by directly specifying priors about the location of the optimum~\cite{Li2020IncorporatingEP,RAMACHANDRAN2020105663,souza2021bayesian,Hvarfner2022}, or structural priors, for example, in the form of log-transformations of hyperparameters~\cite{Hutter2011}, monotonicity constraints~\cite{liaccelerating}, or hyperparameter warping~\cite{Snoek2014}.  

\subsection{Contributions}

The primary goal of this research is to inform the AutoML community and the human-computer-interactions (\ac{HCI}) community about practitioners' motives for using \ac{HPO} methods. By bridging human-centered perspectives from the \ac{HCI} community with technical advancements from the \ac{ML} community, this work makes three key contributions to the development of more effective AutoML tools.

First, we offer a conceptual foundation that outlines why practitioners use \ac{HPO} methods. This foundation comprises six core goals (\eg improving \ac{ML} model performance, aligning with a target audience) and fourteen contextual factors (\eg compute resource availability, traceability of \ac{HPO} methods) that influence choices of practitioners.
From the perspective of \ac{HCI}, the conceptual foundation enables more user-centered \ac{HPO} research by supporting better understanding of practitioner motives. It supports research on human-in-the-loop \ac{ML} by helping define information needs (\eg related to improving transparency in \ac{HPO} tools) and designing better practitioner engagement strategies for different \ac{HPO} methods.
From a technichcal perspective, researchers focusing on \ac{HPO} tool development can leverage these insights to build \ac{HPO} methods and \ac{HPO} tools that extend beyond performance optimization. Future \ac{HPO} tools could be designed to be more context-sensitive, improving their adaptability and utility. Additionally, the identified goals can inform design of benchmarks for evaluating \ac{HPO} tools in terms of compute resource efficiency and automation levels, rather than only \ac{ML} model performance.

Second, we present a mapping between goals, \ac{HPO} methods, and contextual factors, offering insights into why practitioners choose specific \ac{HPO} methods. This mapping helps align \ac{HPO} tool development with real-world practitioner needs.
From the perspective of \ac{HCI}, this work supports better understanding of the decision-making process for \ac{HPO} methods, enabling the design of more intuitive, goal-driven \ac{HPO} tools. This is useful to develop tailored automation features that cater to specific goals and contextual factors.
From a technical perspective, the mapping presents key input parameters (\eg priority of goals and contextual factors) that can be leveraged in development of \ac{HPO} tools. For instance, specialized \ac{HPO} tools could be designed to optimize specific contextual factors rather than using one-size-fits-all approaches.

Third, we present an overview of how practitioners perceive the success of different \ac{HPO} methods in different contexts. The overview highlights areas where existing tools meet expectations and where improvements are needed.
From the perspective of \ac{HCI}, understanding practitioners' perceived success helps inform the design of better decision-support systems for \ac{HPO}. It also enables development of novel interaction concepts that improve how practitioners select and use \ac{HPO} tools.
From a technichcal perspective, by analyzing self-reported success across different goals and contextual factors, we support better understanding of the strengths and weaknesses of \ac{HPO} methods. This can guide the development of new \ac{HPO} methods tailored to specific workflows, practitioner goals, and technical constraints.

\subsection{Limitations and Future Work}

We performed semi-structured interviews in a qualitative and explorative research approach. Interview results strongly rely on interviewees' knowledge, perceptions, and capabilities to verbalize responses to questions---common sources for biases.
We aimed to decrease biases in data gathering and data analysis by reaching out to a variety of practitioners with different levels of experience and work fields. Moreover, we aimed to decrease bias in the analysis of gathered data as multiple analysts independently coded the interview transcripts and discussed their results to agree on a shared understanding. However, despite these efforts, we cannot fully guarantee exhaustiveness and elimination of biases.
Additional goals and contextual factors may be relevant to practitioners. Future research could extend this work to uncover additional goals and contextual factors that were not mentioned by the participants of this study (\eg size of training data sets).

The results presented in this work hint at possible conflicts between goals and contextual factors. Practitioners must resolve such conflicts to succeed in \ac{HPO}, which entails prioritization of goals and the relevance of contextual factors. A possible conflict can arise when practitioners aim to `decrease necessary computations' and `increase \ac{ML} model performance'. Practitioners may attempt to find a Pareto-optimal achievement of both goals based on a clear prioritization.
This work offers a foundation of goals and contextual factors that can lead to trade-offs and call for prioritization of goals. Future work should investigate relationships between goals and contextual factors to uncover such trade-offs and investigate how practitioners resolve them (\eg in terms of prioritization).

Future investigations of human decision-making in \ac{HPO} are a promising research direction that can help to improve \ac{AutoML} by incorporating human knowledge \cite{higuchi2021interactivehpo, wang2019atms, sun2023automl}. The interviewed practitioners reported actions they applied in \ac{HPO}, which are agnostic to the selected \ac{HPO} methods, such as choosing a promising subset of hyperparameters to tune and defining corresponding search ranges.
Most interviewees proceeded very much alike in choosing \ac{HPO} methods, selecting hyperparameters, and tuning hyperparameter configurations. Because those interviewees stated to have successfully attained their goals, such similar proceeding \ac{HPO} allows for the assumption that best practices for actions taken during \ac{HPO} exist. As the interviewees mainly stated that they largely unconsciously compared \ac{HPO} methods but achieved satisfactory outcomes, the identification of cognitive heuristics use in practitioners' decision-making~\cite{Gigerenzer_Brighton_2009,Gigerenzer_Selten_2002} in \ac{HPO} appears to be of great potential to advance automation of \ac{HPO} tasks in AutoML.
By identifying functional heuristics that practitioners use in \ac{HPO}, a better understanding of how practitioners use \ac{HPO} methods can be reached. This could help enhance AutoML, especially \ac{HPO} tools, in terms of automating even complex tasks that are still left to human beings by mimicking human decision-making in a resource-efficient manner. Moreover, such heuristics could be helpful to make \ac{HPO} tools more efficient.

\section{Conclusion}
\label{sec:conclusion}

While programmatic \ac{HPO} methods, such as Bayesian optimization and evolutionary algorithms, achieve high efficiency of the \ac{HPO} process; practitioners sometimes opt for efficiency-wise inferior \ac{HPO} methods, such as grid search and manual tuning.
To understand practitioner motives for using \ac{HPO} methods, we performed a two-step research approach consisting of semi-structured interviews and a survey based on an online questionnaire.
We identified six principal goals pursued by practitioners in \ac{HPO}, such as\textit{ decrease practitioner effort}, \textit{decrease necessary computations}, and \textit{increase \ac{ML} model understanding}.
Moreover, we extracted fourteen contextual factors that influence practitioners' decisions for using \ac{HPO} methods, such as \textit{available compute resources}, \textit{\ac{HPO} method traceability}, and \textit{parallelization possibilities}.

By bridging the gap between technological advancements and practitioner motives to use \ac{HPO} methods, this work contributes to enhancing \ac{HPO} practices and tools in the context of AutoML.
In particular, the results of this study can guide development of more user-centered \ac{HPO} methods and \ac{HPO} tools that cater to practitioner motives.

This work calls for more user-centered research on \ac{HPO}, particularly on exploring purposeful ways to involve practitioners in programmatic \ac{HPO} methods, decision support systems for \ac{HPO}, and enhancing transparency and comprehensibility of programmatic \ac{HPO} methods.
We will build on the findings presented in this work and seek to identify functional human heuristics \cite{Gigerenzer_Brighton_2009} applied in \ac{HPO}. After identifying human heuristics (\eg \cite{godbole-github23}), we aim to implement them in algorithms for \ac{AutoML} and evaluate those algorithms in comparison to the performance of human decision-making and less human-centered \ac{HPO} tools.

\begin{acks}
We thank all study participants for their time and valuable contributions, which formed the foundation for this work. In particular, we thank Mikael Beyene and Benjamin Sturm for their fruitful input. This work was supported by KASTEL Security Research Labs.
Frank Hutter and Marius Lindauer acknowledge funding by the European Union (via ERC Consolidator Grant DeepLearning 2.0, grant no.~101045765, and ERC Starting Grant ``ixAutoML'', grant no.~101041029, respectively). 
The views and opinions expressed are those of the author(s) only and do not necessarily reflect those of the European Union or the European Research Council. Neither the European Union nor the granting authority can be held responsible for them.
\begin{center}\includegraphics[width=0.3\textwidth]{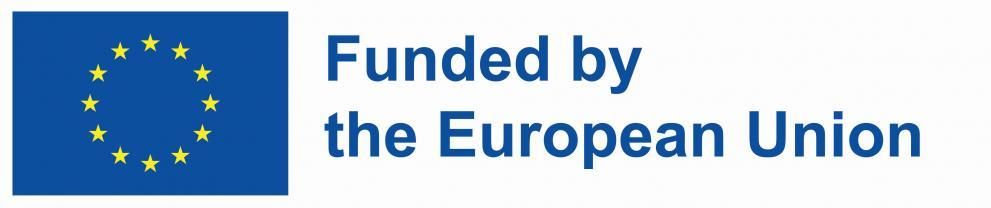}\end{center}
\end{acks}

\newpage
\bibliographystyle{ACM-Reference-Format}
\bibliography{acmart}

\newpage
\appendix
\section*{Appendix}
\section{Interview Guide}

The following presents the interview guide that we used in the semi-structured interviews.

\subsection{Briefing}
Dear study participant,\\

\noindent Thank you for supporting our research on hyperparameter optimization (HPO) in machine learning (ML) with your time and expertise.
The interview will take about 30 minutes of your time. All data collected will be treated confidentially and reported only in aggregated form. Your responses will not be linked to your identity in any future publications.

HPO is an increasingly important topic in ML research, as it helps enhance model performance and enables objective comparisons of ML methods. In this study, we define HPO as the process of iteratively improving hyperparameter configurations during ML model training. A hyperparameter is a parameter whose value controls the learning process.

Various tools and ML libraries (e.g., Keras Tuner) support automated hyperparameter optimization. However, it remains unclear why practitioners often do not use advanced automated HPO\footnote{Xavier Bouthillier, Gaël Varoquaux. Survey of machine learning experimental methods at NeurIPS2019 and ICLR2020. [Research Report] Inria Saclay Ile de France. 2020. \url{https://hal.science/hal-02447823/document}}, and in which contexts manual HPO may even outperform automated optimization and vice versa.
Our goal in this study is to understand the motives of practitioners for choosing different HPO methods.\\

All questions in this interview relate to your personal experience in ML, including the HPO methods you have used for hyperparameter optimization. To obtain detailed information on your work and its context, we ask you to refer to one ML model you developed for productive use in academia or industry. If possible, please provide a link to the publication or associated repository (e.g., GitHub).

The remainder of the interview is structured into three sections:
\begin{enumerate}
    \item \textbf{Hyperparameter Optimization in Machine Learning:} We will discuss the ML project you will reference in your responses and seek to understand your choice of HPO methods.
    \item \textbf{Participant Background and Personal Experiences:} We will ask about your demographics, background, and expertise in HPO and ML.
    \item \textbf{Debriefing:} We will summarize the key findings from the interview and outline the next steps.
\end{enumerate}

While we have prepared several questions, we welcome any additional insights you wish to share.\\

We appreciate your participation and encourage you to share your experiences and perspectives on HPO approaches. There are no wrong answers.\\

Thank you again for your time and participation!

\newpage


\subsection{Hyperparameter Optimization in Machine Learning}

This section is divided into two subsections. First, we will ask you to select one of your previous ML projects to focus on during the interview and to describe the HPO methods you used. Second, we will ask about your reasons for choosing these methods.

\subsubsection{Hyperparameter Optimization in a Selected Machine Learning Project}

\paragraph{Details on Machine Learning Model}

In the interview, we would like to refer to one of your previous ML projects.
Please select an ML project that was meant for productive use in academia or industry, and where you optimized hyperparameters.

\begin{enumerate}
    \item \textit{Optional:} To which repository refer your following answers? Please provide the link to the repository of your model. \\
    URI: \underline{\hspace{10cm}}
    \item In case your repository includes multiple models: please provide information to which exact ML model your following answers will refer. \\
    \underline{\hspace{\linewidth}}
    \item Was you ML model empirically evaluated (e.g., in the form of a benchmark)?\\
    $\Box$ Yes \quad $\Box$ No\\
\end{enumerate}

\paragraph{Optimization Approach}
Next, we want to learn more about how you tuned hyperparameters of the selected ML model.\\

\begin{mdframed}[backgroundcolor=black!10]
\textbf{Interviewer Instruction:} 
Question~\ref{item:did-you-use-hpo} must be answered `Yes'. Otherwise, the interviewee must not participate in the study.
\end{mdframed}

\vspace{5pt}

\begin{enumerate}\addtocounter{enumi}{3}
    \item\label{item:did-you-use-hpo} Did you optimize the hyperparameters of the model selected for this interview?\\
    $\Box$ Yes \quad $\Box$ No\\
    \item How did you tune your hyperparameters? If you have used a combination of different HPO methods, name all methods used.\\
    \textit{Exemplary answers are Bayesian optimization, grid search, manual tuning, and random search.}\\
    \underline{\hspace{\linewidth}} \\
\end{enumerate}

\newpage

\subsubsection{Motives to Use the Selected Hyperparameter Optimization Methods}
\label{sec:motive}

\begin{mdframed}[backgroundcolor=black!10]
\textbf{Interviewer Instruction:} 
This section builds on the responses from the previous section. The used HPO methods determine whether the questions need to be inverted by using `\textit{(not)}', as indicated in the following questions.
\end{mdframed}

\vspace{5pt}

In the previous section, you indicated the HPO methods you used for your ML model:
\texttt{[LIST OF USED HPO METHODS]}.
We are now interested in your reasons for choosing these methods, as well as the advantages and disadvantages you experienced. We value your personal experiences and will not judge your responses---there are no wrong answers.

\vspace{5pt}

\begin{enumerate}\addtocounter{enumi}{5}
    \item Which goals (e.g., increase ML model comprehension) did you try to reach through HPO by the chosen methods?
    A goal is an idea of the future or desired results where the achievement of the idea is decidable. If multiple HPO methods have been used, please explain in which order you used them and explain the individual goals you tried to achieve. \\
    \underline{\hspace{\linewidth}} \\
    \underline{\hspace{\linewidth}} \\
    \underline{\hspace{\linewidth}}
    \item What potential advantages and disadvantages regarding the achievement of the described goals are you aware of with respect to the (combination of) HPO methods you have chosen? What potential advantages and disadvantages did you encounter in achieving these goals with the selected HPO methods? If you used multiple HPO methods, please specify which method each advantage and disadvantage refers to. \\
    \underline{\hspace{\linewidth}} \\
    \underline{\hspace{\linewidth}} \\
    \underline{\hspace{\linewidth}}
    \item Why did you \textit{(not)} choose random search? Please consider contextual factors (e.g., limited compute) and goals (e.g., improve model comprehension). \\
    \underline{\hspace{\linewidth}} \\
    \underline{\hspace{\linewidth}} \\
    \underline{\hspace{\linewidth}}
    \item Why did you \textit{(not)} manual tuning for HPO? Please consider contextual factors (e.g., limited compute) and goals (e.g., improve model comprehension). \\
    \underline{\hspace{\linewidth}} \\
    \underline{\hspace{\linewidth}} \\
    \underline{\hspace{\linewidth}}
    \item Why did you \textit{(not)} grid search for HPO? Please consider contextual factors (e.g., limited compute) and goals (e.g., improve model comprehension). \\
    \underline{\hspace{\linewidth}} \\
    \underline{\hspace{\linewidth}} \\
    \underline{\hspace{\linewidth}}
    \item Why did you \textit{(not)} choose Bayesian optimization for HPO? Please consider contextual factors (e.g., limited compute) and goals (e.g., improve model comprehension). \\
    \underline{\hspace{\linewidth}} \\
    \underline{\hspace{\linewidth}} \\
    \underline{\hspace{\linewidth}}
    \item Which tools did you use for HPO? Exemplary tools are HPBandster, hyperopt, Ray, spearmint, and Ax. \\
    \underline{\hspace{\linewidth}} \\
    \underline{\hspace{\linewidth}} \\
    \underline{\hspace{\linewidth}}
    \item How could future research support you in optimizing hyperparameters? \\
    \underline{\hspace{\linewidth}} \\
    \underline{\hspace{\linewidth}} \\
    \underline{\hspace{\linewidth}}
    \item After you have helped us with your expertise, we would be grateful if you would share your thoughts on the contexts in which automated HPO may outperform manual HPO and vice versa. \\
    \underline{\hspace{\linewidth}} \\
    \underline{\hspace{\linewidth}} \\
    \underline{\hspace{\linewidth}}
\end{enumerate}

\subsection{Participant Background and Personal Experiences}
In this section, we will ask you questions about your experience in artificial intelligence to better understand the context of your previous answers. Your personal information will not be disclosed in any way that could identify you.

\begin{enumerate}\addtocounter{enumi}{14}
    \item What area(s) of artificial intelligence do you focus on?\\
    \textit{Exemplary answers are automatic speech recognition, computer vision, and natural language processing.}\\
    \underline{\hspace{\linewidth}}
    
    \item What is your main activity in artificial intelligence?\\
    \textit{Exemplary answers are consulting, development of statistical methods, library development, software solution development (i.e., using existing libraries and methods), and use case development.}\\
    \underline{\hspace{\linewidth}}
        
    \item What is/are your area(s) of expertise? \\
    \textit{Exemplary answers are AutoML, parallel computing, or (un-)supervised learning.}\\
    \underline{\hspace{\linewidth}}
    
    \item In which field do you work?\\
    \textit{Exemplary answers are academia, automotive, finance, IT support \& services, and pharma.}\\
    \underline{\hspace{\linewidth}}
    
    \item How many years of experience do you have in ML?\\
    $\Box$ \textless 2 \quad $\Box$ 2-4 \quad $\Box$ 5-7 \quad  $\Box$ 8-10 \quad $\Box$ 11-15 \quad $\Box$ \textgreater 15\\
        
    \item What is your highest educational level?\\
    \textit{Exemplary answers are Bachelor of Engineering, Bachelor of Science, Master of Science, and PhD.}\\
    \underline{\hspace{\linewidth}}
    
    \item What is your profession?\\
    \textit{Exemplary answers are big data engineer, developer, data scientist, and ML engineer.}\\
    \underline{\hspace{\linewidth}}
        
    \item What is your position in your organization?\\ \textit{Exemplary answers are professor, PhD student, or lead software architect.} \\
    \underline{\hspace{\linewidth}} \\
    
    \item What is your age?\\
        $\Box$  \textless 20 \quad $\Box$  20-25 \quad $\Box$ 26-30 \quad $\Box$ 31-35 \quad $\Box$ 36-40 \quad $\Box$ 41-45
        $\Box$ 46-50 \quad $\Box$ 51-55 \quad $\Box$ 56-60 \quad $\Box$ 61-65 \quad $\Box$ $\textgreater 65$\\
            
    \item In which country are you primarily working or employed? \\
    \underline{\hspace{\linewidth}} \\
    
    \item How many people are employed at your organization?\\
    $\Box$ \textless 10 \quad $\Box$ 10-50 \quad $\Box$ 51-150 \quad $\Box$ 150-500 \quad $\Box$ 501-1000 \quad $\Box$  1,000\\
    
\end{enumerate}

\subsection{Debriefing}
\label{sec:debriefing}

This is the final section of the interview. Thank you again for supporting our research with your valuable time and expertise.

\vspace{5pt}

\begin{mdframed}[backgroundcolor=black!10]
\textbf{Interviewer Instruction:} 
Please summarize the key findings from the interview.
\end{mdframed}

\vspace{5pt}

\begin{enumerate}\addtocounter{enumi}{25}
    \item Would you like to receive a summary of the study results?
    \item Do you know colleagues who might be interested in participating in this study? If yes, could you please connect us with them?
\end{enumerate}

\vspace{5pt}

\begin{mdframed}[backgroundcolor=black!10]
\textbf{Interviewer Instruction:} 
 Outline the next steps.
\end{mdframed}

\vspace{5pt}

We have now reached the end of the interview. Thank you very much for your participation and support!

\newpage


\end{document}